\documentclass[11pt]{article}

\usepackage{acl}

\usepackage{times}
\usepackage{latexsym}
\usepackage[T1]{fontenc}
\usepackage[utf8]{inputenc}
\usepackage{microtype}

\usepackage{graphicx}
\usepackage{amsmath,amssymb}
\usepackage{booktabs}
\usepackage{algorithm}
\usepackage{algorithmicx}
\usepackage{algpseudocode}
\usepackage{multirow}
\usepackage[table]{xcolor}
\usepackage{listings}
\definecolor{bestblue}{RGB}{204,229,255}
\definecolor{bestgreen}{RGB}{204,255,204}

\title{From Evidence-Based Medicine to Knowledge Graph: Retrieval-Augmented Generation for Sports Rehabilitation and a Domain Benchmark}

\author{
Jinning Zhang$^{1,2*}$ \quad
Jie Song$^{1,2*}$ \quad
Wenhui Tu$^{1,2\ddagger}$ \quad
Zecheng Li$^{1,2\ddagger}$ \\
\textbf{
Jingxuan Li$^{3}$ \quad
Jin Li$^{4}$ \quad
Xuan Liu$^{5}$ \quad
Taole Sha$^{7}$ \quad
Zichen Wei$^{6}$ \quad
Yan Li$^{1,2\dag}$
} \\[6pt]
$^{1}$Beijing Key Laboratory of Sports Performance and Skill Assessment, \\
Beijing Sport University, Beijing 100084, China \\
$^{2}$Department of Exercise Physiology, School of Sport Science, \\
Beijing Sport University, Beijing 100084, China \\
$^{3}$School of Sport Medicine and Physical Therapy, \\
Beijing Sport University, Beijing 100084, China \\
$^{4}$Rehabilitation Center, Beijing Rehabilitation Hospital, Beijing 100144, China \\
$^{5}$Optum Care Washington, Everett, Washington 98201, USA \\
$^{6}$School of Management, Beijing Sport University, Beijing 100084, China \\
$^{7}$Department of Statistics and Actuarial Science, \\
The University of Hong Kong, Hong Kong, China \\[4pt]
{\small $^{*}$Equal contribution \quad $^{\ddagger}$Equal contribution as second authors \quad $^{\dag}$Corresponding author: \texttt{bsuliyan@bsu.edu.cn}}
}

\begin{document}
\maketitle

\begin{abstract}
Current medical retrieval-augmented generation (RAG) approaches overlook evidence-based medicine (EBM) principles, leading to two key gaps: (1) the lack of PICO alignment between queries and retrieved evidence, and (2) the absence of evidence hierarchy considerations during reranking. We present SR-RAG, an EBM-adapted GraphRAG framework that integrates the PICO framework into knowledge graph construction and retrieval, and proposes Bayesian Evidence Tier Reranking (BETR) to calibrate ranking scores by evidence grade without predefined weights. Validated in sports rehabilitation, we release a knowledge graph (357,844 nodes, 371,226 edges) and a benchmark of 1,637 QA pairs. SR-RAG achieves 0.812 evidence recall@10, 0.830 nugget coverage, 0.819 answer faithfulness, 0.882 semantic similarity, and 0.788 PICOT match accuracy, substantially outperforming five baselines. Five expert clinicians rated the system 4.66--4.84 on a 5-point Likert scale, and system rankings are preserved on a human-verified gold subset ($n=80$).
\end{abstract}

\section{Introduction}

Consider the following question posed to a large language model (LLM): ``My child has congenital heart disease and has just undergone surgery. How should we conduct postoperative exercise rehabilitation?'' As of 2024, the evidence base for rehabilitation in children with congenital heart disease remains limited, consisting primarily of observational studies \citep{barbazi_exploring_2025,ubeda_tikkanen_core_2023}, with authoritative guidelines only recently emerging \citep{noauthor_2024_2024}. In such cases, LLMs often provide outdated or suboptimal answers due to their training data cutoff \citep{hager_evaluation_2024}.

In medicine, clinical evidence evolves rapidly. For example, the latest American College of Sports Medicine (ACSM) edition adopts the metabolic chronotropic reserve (MCR) to assess whether the heart rate response during exercise is appropriate \citep{acsm_guidelines_2025}, replacing the traditional 220 minus age formula, which recent studies have shown can carry significant errors \citep{almaadawy_target_2024,lauer_impaired_1999}. Retrieval-Augmented Generation (RAG) is designed to incorporate such time-sensitive information, dynamically extending LLM knowledge to prevent obsolescence and reduce hallucinations \citep{lewis_retrieval-augmented_2020}. In medicine, the need for traceable evidence and timely information makes RAG a widely adopted approach for enhancing LLM reliability \citep{yang_rag_healthcare_2025,bechard_reducing_2024}.

Returning to the original user question, the patient is a child with congenital heart disease---a special population. Recently evolved RAG frameworks typically focus on performance improvements without explicitly accounting for query-specific target populations \citep{gupta_rag_survey_2024,xiong_mirage_2024,asai_selfrag_2024,yan_crag_2024}, potentially generating population-mismatched answers. Crucially, most evaluation metrics cannot identify population mismatches. For example, semantic similarity uses cosine similarity to quantify differences between generated and reference answers at a macro level, making general phrasing far more influential than specific details; answer faithfulness measures only factual alignment between the answer and retrieved context, ignoring relevance to the query \citep{es_ragas_2024}. Another key issue is that even if RAG incorporates the latest authoritative guidelines for the rehabilitation of children with congenital heart disease, these sources would still be weighted the same as earlier scattered observational evidence during retrieval \citep{sun_metarag_2025}, which violates the concept of a hierarchy of evidence \citep{guyatt_grade_2008}. The two issues above can be summarized as follows: Current medical RAG approaches neglect the evidence-based medicine (EBM) framework \citep{sackett_ebm_1996}. In particular, the Population--Intervention--Comparator--Outcome (PICO) framework and the hierarchy of evidence remain largely overlooked.

Regarding the combination of EBM and RAG, especially the integration of the PICO framework, graph-based retrieval-augmented generation (GraphRAG) shows promising potential. This paradigm leverages knowledge graphs constructed from entity-relation-attribute triples to provide a structured hierarchy for the corpus \citep{edge_graphrag_2024}. Originally, it is designed to enhance the multihop reasoning capabilities of LLMs for complex questions, which are particularly important in medical contexts \citep{cabello_meg_2024}. However, this highly structured arrangement can also implicitly guide LLMs to retrieve from nodes aligned with the query, thereby potentially improving PICO alignment between questions and answers. Based on this, we propose the key question of this study:

\begin{itemize}
\item How can the EBM framework be adapted to RAG pipelines, particularly GraphRAG?
\end{itemize}

\begin{figure}[t]
\centering
\includegraphics[width=\columnwidth]{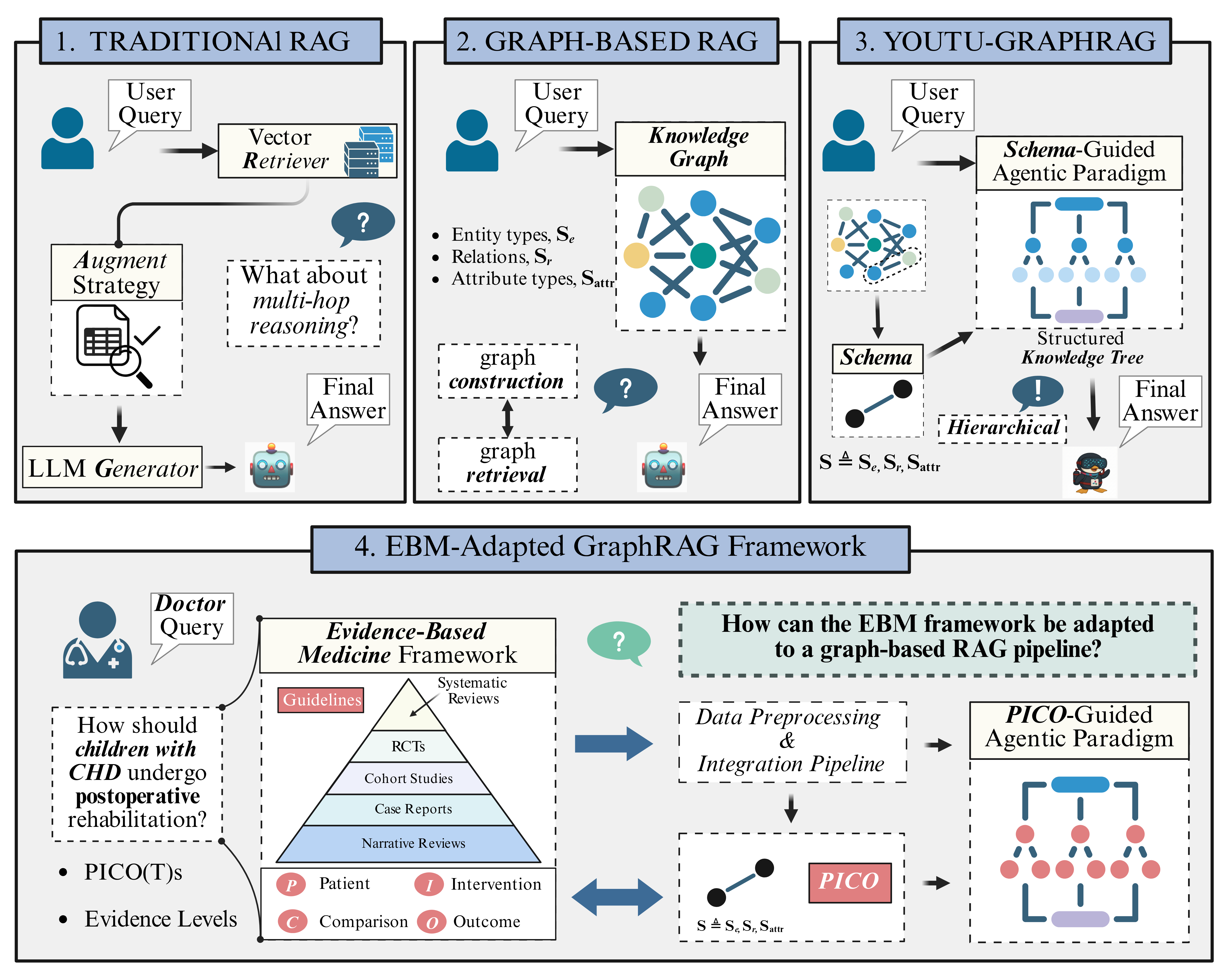}
\caption{Evolution from traditional RAG to graph-based RAG and the proposed EBM-adapted GraphRAG framework integrating evidence hierarchy and the PICO framework.}
\label{fig:framework}
\end{figure}

Since the pioneering work of \citet{edge_graphrag_2024}, GraphRAG has evolved along two main trajectories: graph retrieval and graph construction \citep{dong_youtu_graphrag_2025}. The former focuses on improving retrieval efficiency and relevance, while the latter aims to enrich graph topology and enable multi-granularity retrieval. Recently, Youtu-GraphRAG, proposed by \citet{dong_youtu_graphrag_2025}, unifies graph retrieval and graph construction through schema-bounded agentic extraction. This strategy reduces extraction noise during graph construction while enabling rapid domain adaptation through schema replacement. Essentially, it shifts the implicit retrieval constraints of traditional GraphRAG to a new paradigm of explicitly constrained retrieval by predefining the schema. In the medical domain, when we directly define the schema as PICO-related entity types, PICO mismatch between queries and retrieved chunks may be substantially reduced (Fig.~\ref{fig:framework}). Accordingly, this study aims to address EBM adaptation based on the Youtu-GraphRAG framework. In particular, although the field of sports rehabilitation is resource-rich, no domain-specific RAG system or reusable benchmark currently exists. We therefore focus on this domain, curating a corpus, constructing a benchmark, and validating our framework.

In summary, the main contributions of this study are as follows:

\begin{itemize}
\item We curate a sports rehabilitation corpus (21 conditions) and construct a knowledge graph comprising 357,844 nodes and 371,226 edges.
\item We propose BETR, a Bayesian-inspired reranking algorithm that calibrates ranking scores by learning evidence-grade biases from data without predefined weights.
\item We present SR-RAG, integrating the EBM framework into GraphRAG through a generalizable adaptation strategy.
\item We develop a reusable benchmark (1,637 QA pairs) and validate SR-RAG through both automated metrics and expert clinician review.
\end{itemize}

\section{Related Work}

\subsection{Graph-Based RAG}

The original GraphRAG \citep{edge_graphrag_2024} relies on community detection to produce hierarchical summaries for global queries, but its fixed community granularity limits recall for fine-grained medical questions. LightRAG addresses this with dual-level retrieval operating at both entity and relationship granularities \citep{guo_lightrag_2024}. HippoRAG takes a different approach by mimicking hippocampal memory indexing with personalized PageRank \citep{gutierrez_hipporag_2024}. Youtu-GraphRAG further introduces dually-perceived community compression that jointly considers relational and semantic affinity, producing a four-layer knowledge tree that supports multi-granularity retrieval \citep{dong_youtu_graphrag_2025}. Despite these advances, all existing GraphRAG systems treat the extraction schema as a generic tool; none has tailored it to domain-specific clinical structures such as PICO for medical retrieval.

\subsection{EBM-Aligned RAG}

Recent RAG studies in medicine have begun to emphasize integration with EBM. PICOs-RAG improves PICO alignment by expanding and standardizing user questions and extracting PICO elements for retrieval \citep{sun_picos_rag_2025}. Its key design is query rewriting rather than hypothetical document generation; however, a semantic gap typically exists between reformulated queries and retrieved documents \citep{wang_query2doc_2023,zhu_hyqe_2024}. Similarly, Quicker applies the PICO principle across question decomposition, document retrieval, and study screening, and implements the GRADE system to evaluate evidence quality across five dimensions \citep{wang_quicker_2025,guyatt_grade_2008}.

Regarding evidence hierarchy, Med-R$^2$ introduces a dedicated reranking module that maps evidence hierarchy to confidence levels via a manually defined linear mapping \citep{wang_medr2_2025}. META-RAG further refines this by drawing on meta-analysis principles to rerank evidence across reliability, heterogeneity, and extrapolation, introducing the DerSimonian--Laird random-effects model for heterogeneity analysis \citep{sun_metarag_2025}. However, both systems rely on manually defined score formulas whose weight formulations lack theoretical justification and cannot adapt to evidence distributions across domains.

These works demonstrate growing interest in EBM-aligned RAG but leave two gaps: (1) no existing system integrates evidence hierarchy into the reranking objective through a principled learning framework, and (2) the PICO structure has not been exploited at the graph schema level for retrieval.

\section{Method}

\subsection{Corpus Collection and Data Preprocessing}

We retrieved open-access and institution-subscribed literature from PubMed and Embase, supplemented by authoritative sports rehabilitation organization websites. Referring to established evidence grading frameworks \citep{ocebm_levels_2011,nhmrc_evidence_2009,jbi_levels_2014}, each document was assigned an evidence grade through manual abstract review: A = guidelines and expert consensus, B = systematic reviews and meta-analyses, C = RCTs, D = cohort studies, E = other research. The final corpus covers 21 common sports rehabilitation conditions. PDFs were converted to Markdown via Docling \citep{auer_docling_2024}, with evidence grades recorded in metadata for downstream retrieval and BETR calibration. For chunking, we adopted an LLM-aware hybrid strategy: documents were first split by heading structure, then the LLM performed semantic grouping of atomic blocks into evidence windows \citep{jain_autochunker_2025,zhao_metachunking_2024,duarte_lumberchunker_2024}; details are provided in Appendix~A.

\subsection{BETR Algorithm}

The evidence hierarchy principle in EBM was integrated into the reranking pipeline to optimize ranking order. Existing evidence-hierarchy-based reranking paradigms mostly rely on subjective preset scores, yielding heuristic weighting schemes \citep{wang_medr2_2025}. To address this, we proposed BETR, which introduces evidence hierarchy as an ordered structure into the ranking calibrator via a data-driven paradigm. The complete workflow is presented in Algorithm~\ref{alg:betr}.

\begin{algorithm*}[t]
\caption{Bayesian Evidence Tier Reranking}
\label{alg:betr}
\small
\begin{algorithmic}[1]
\Require Disjoint query splits $\mathcal{Q}_{\text{train}},\mathcal{Q}_{\text{val}}$ with gold windows $\mathcal{W}^\star(q)$;
candidate generator $\mathrm{Cand}(\cdot)$;
reranker $f_\theta$ returning logit $s(q,d)$;
ordered evidence grades $A \succ B \succ C \succ D \succ E$;
grade function $\mathrm{Grade}(d)\in\{A,B,C,D,E\}$;
negatives per positive $K$;
shrinkage scale $\tau$ (selected via grid search on $\mathcal{Q}_{\text{val}}$ and fixed for all experiments);
scale prior $\sigma_a$ for $a$.
\Ensure Calibrator parameters $(\alpha,\delta_B,\delta_C,\delta_D,\delta_E)$ and online ranking score $r(q,d)$.

\Statex \textbf{Step 1: Build pairwise records (train split)}
\State $\{\mathcal{P}_q\}_{q\in\mathcal{Q}_{\text{train}}}\gets\emptyset$
\For{$q\in\mathcal{Q}_{\text{train}}$}
    \State $\mathcal{C}_q \gets \mathrm{Cand}(q)$;\quad $\mathcal{C}_q^+ \gets \mathcal{C}_q \cap \mathcal{W}^\star(q)$;\quad $\mathcal{C}_q^- \gets \mathcal{C}_q \setminus \mathcal{W}^\star(q)$
    \State If $\mathcal{C}_q^+=\emptyset$ or $\mathcal{C}_q^-=\emptyset$, set $\mathcal{P}_q\gets\emptyset$.
    \State Else form $\mathcal{P}_q\subseteq \mathcal{C}_q^+\times \mathcal{C}_q^-$ by sampling $K$ negatives per $d^+\in \mathcal{C}_q^+$.
    \State For each $(d^+,d^-)\in\mathcal{P}_q$, compute $\Delta s=s(q,d^+)-s(q,d^-)$ and $t^\pm=\mathrm{Grade}(d^\pm)$.
\EndFor

\Statex \textbf{Step 2: Ordered grade effects and MAP fit}
\Statex Define $u_A=0$, $u_B=-\delta_B$, $u_C=-(\delta_B+\delta_C)$, $u_D=-(\delta_B+\delta_C+\delta_D)$, $u_E=-(\delta_B+\delta_C+\delta_D+\delta_E)$, with $\delta_B,\delta_C,\delta_D,\delta_E\ge 0$.
\Statex Reparameterize $a=\exp(\alpha)$.
\Statex Fit $(\alpha,\delta_B,\delta_C,\delta_D,\delta_E)$ by maximizing the query-normalized MAP objective:
\begin{equation*}
\begin{aligned}
\max_{\alpha,\delta_B,\delta_C,\delta_D,\delta_E\ge 0}\quad
& \frac{1}{|\mathcal{Q}_{\text{train}}|}\sum_{q\in\mathcal{Q}_{\text{train}}}
\frac{1}{\max(1,|\mathcal{P}_q|)}\sum_{(d^+,d^-)\in\mathcal{P}_q}
\log \sigma(a\,\Delta s + u_{t^{+}}-u_{t^{-}}) \\
&\quad -\frac{1}{2\tau^2}\big(\delta_B^2+\delta_C^2+\delta_D^2+\delta_E^2\big)
-\frac{1}{2\sigma_a^2}\alpha^2.
\end{aligned}
\end{equation*}
\Statex where $\sigma(z)=(1+e^{-z})^{-1}$.

\Statex \textbf{Step 3: Online ranking}
\State For a new query $q$ and each candidate window $d\in \mathrm{Cand}(q)$, set $a=\exp(\alpha)$ and $t=\mathrm{Grade}(d)$.
\State Compute $r(q,d)=a\,s(q,d)+u_t$ and rank by $r(q,d)$.
\end{algorithmic}
\end{algorithm*}

\paragraph{Task Definition.} Given a clinical question $q$ and a candidate evidence window $d$, the reranker outputs an uncalibrated relevance score $s(q,d)$. We aimed to combine $s(q,d)$ with evidence hierarchy $\mathrm{Grade}(d) \in \{A \succ B \succ C \succ D \succ E\}$ to yield a unified ranking score $r(q,d)$ satisfying: (1) when semantic relevance differences are large, $s(q,d)$ dominates the ranking; (2) when candidates have comparable $s(q,d)$, higher evidence grades receive higher final scores. The five evidence grades follow the hierarchy defined in \S3.1.

\paragraph{Training Labels.} For each question $q$, candidate windows included in the reference evidence chain serve as positive examples: $\mathcal{C}_q^+=\mathcal{C}_q\cap\mathcal{W}^\star(q)$; the remainder serve as negatives: $\mathcal{C}_q^-=\mathcal{C}_q\setminus\mathcal{W}^\star(q)$. This design avoids per-item manual annotation and aligns the ranking objective with evidence window selection. $\mathcal{W}^\star(q)$ is constructed via independent relevance grading over the full candidate pool (\S3.5) and typically spans multiple evidence grades.

\paragraph{Pairwise Ranking Objective.} BETR adopts a pairwise learning-to-rank approach \citep{burges_ranknet_2005}, learning preferences through pairwise comparisons. Specifically, for the candidate pool of question $q$, we construct positive--negative pairs $(d^+, d^-)$ and model preference probability via the Bradley--Terry model \citep{bradley_terry_1952}:
\begin{equation}
P(d^+ \succ d^- \mid q)=\sigma\big(a\Delta s + u_{t^+}-u_{t^-}\big)
\end{equation}
where $\Delta s = s(q,d^+) - s(q,d^-)$ is the semantic relevance difference; $t^+=\mathrm{Grade}(d^+)$ and $t^-=\mathrm{Grade}(d^-)$ denote evidence grades; $u_t$ is the grade bias; $a > 0$ is the scale parameter; and $\sigma(\cdot)$ is the sigmoid function. This formulation jointly considers two signals: (1) semantic relevance difference $\Delta s$, and (2) evidence grade difference $u_{t^+} - u_{t^-}$. When semantic scores are comparable, windows with higher evidence grades receive additional positive bias.

\paragraph{Ordered Hierarchical Parameterization.} To explicitly encode a pyramid-shaped evidence hierarchy and prevent grade inversion under noisy labels, we fix the grade ordering as $A\succ B\succ C\succ D\succ E$ and adopt monotonically constrained incremental parameterization:
\begin{equation}
\begin{aligned}
u_A &= 0,\quad u_B = -\delta_B, \\
u_C &= -(\delta_B+\delta_C),\quad \ldots, \\
u_E &= -(\delta_B+\delta_C+\delta_D+\delta_E)
\end{aligned}
\end{equation}
where $\delta_B,\delta_C,\delta_D,\delta_E\ge 0$, naturally guaranteeing $u_A\ge u_B\ge u_C\ge u_D\ge u_E$, yielding an evidence-grade pyramid consistent with EBM. We then cast BETR parameter estimation in a Bayesian framework, jointly learning the scale parameter $a$ and grade increments $\boldsymbol{\delta}=(\delta_B,\delta_C,\delta_D,\delta_E)$ via MAP estimation.

\paragraph{MAP Estimation.} We impose zero-centered Gaussian priors on the parameters, encoding the default assumption of no grade bias: $\alpha \sim \mathcal{N}(0, \sigma_a^2)$, $\delta_i \sim \mathcal{N}^+(0, \tau^2)$, where $\mathcal{N}^+$ denotes a Gaussian truncated to nonnegative values. This prior has maximum density at $\delta=0$, implying that ranking defaults to being driven by semantic relevance $s(q,d)$; the prior for $\alpha$ is centered at $0$, corresponding to $a=\exp(\alpha)\approx 1$, so that semantic scores and grade biases are summed on the same scale. We then fit $(\alpha, \delta_B, \delta_C, \delta_D, \delta_E)$ via maximum a posteriori estimation (full derivation in Appendix~B):
\begin{equation}
\label{eq:map}
\begin{aligned}
\max_{\alpha,\;\boldsymbol{\delta} \ge \mathbf{0}}\; & \frac{1}{|\mathcal{Q}_{\mathrm{train}}|}\sum_{q\in\mathcal{Q}_{\mathrm{train}}}\frac{1}{\max(1,|\mathcal{P}_q|)}\sum_{(d^+,d^-)\in\mathcal{P}_q} \\
& \log \sigma\!\big(a\,\Delta s + u_{t^+} - u_{t^-}\big) \\
& - \frac{1}{2\tau^2}\big(\delta_B^2+\delta_C^2+\delta_D^2+\delta_E^2\big) - \frac{1}{2\sigma_a^2}\alpha^2
\end{aligned}
\end{equation}
where $a=\exp(\alpha)$. The first term is the query-normalized pairwise log-likelihood; the remaining terms are quadratic shrinkage penalties. This framework ensures that the posterior is likelihood-dominated when data are ample and shrinks toward the prior when data are scarce, achieving adaptive regularization. The hyperparameter $\tau$ is selected via grid search on a validation set and fixed across all experiments. At inference, the final ranking score is $r(q,d)=\hat{a}\,s(q,d)+\hat{u}_{\mathrm{Grade}(d)}$ .

\subsection{PICO-extended Schema and Knowledge Graph Construction}

During knowledge graph construction, we instantiated the Youtu-GraphRAG seed schema as a PICO-extended schema and followed its graph construction workflow to build the knowledge graph over the full corpus.

\paragraph{Schema Definition and Constrained Extraction.} Following Youtu-GraphRAG, the schema is defined as $\mathcal{S}\triangleq\langle \mathcal{S}_e,\mathcal{S}_r,\mathcal{S}_{\text{attr}}\rangle$, specifying entity, relation, and attribute types. The LLM extraction agent performs constrained triple extraction: all extracted entity--relation--entity triples and entity--attribute pairs must map to types in $\mathcal{S}$, suppressing out-of-schema noise.

\paragraph{PICO Instantiation.} We instantiate $\mathcal{S}$ as PICO-related types: (1) entity types $\mathcal{S}_e^{\text{PICO}}$: Population, Condition, Intervention, Comparator, Outcome, Timepoint, plus domain-specific extensions (e.g., Arm, Device); (2) relation types $\mathcal{S}_r^{\text{PICO}}$: directed relations linking studies to PICO entities (e.g., has\_population, uses\_intervention, reports\_outcome); (3) attribute types $\mathcal{S}_{\text{attr}}^{\text{PICO}}$: key dimensions refining PICO elements (e.g., age\_bin, followup\_weeks, protocol\_params). A complete list is provided in Appendix~C. The community compression algorithm and knowledge tree indexing follow Youtu-GraphRAG without modification (Appendix~C).

\subsection{SR-RAG Pipeline}

SR-RAG introduces three key improvements to Youtu-GraphRAG: (1) a PICO-guided HyDE channel fused with graph retrieval via RRF; (2) two-stage reranking using ColBERT and a cross-encoder; (3) evidence-grade-aware retrieval followed by BETR-based reranking.

\paragraph{PICO-guided HyDE.} HyDE generates hypothetical documents to bridge the semantic gap between queries and corpus \citep{gao_hyde_2023,wang_query2doc_2023}. We incorporated PICO soft constraints into the HyDE prompt: available P/I/C/O/T keywords were extracted from the query as anchors (missing fields are permitted). Hypothetical documents must reuse these anchors and are prohibited from fabricating missing fields. This ensures HyDE serves purely as a retrieval intermediary for semantic alignment, without deviating from the original query's PICO elements.

\paragraph{Two-stage Reranking.} We first used ColBERT (mxbai-edge-colbert-v0) as the coarse ranking model \citep{takehi_mxbai_edge_2025}. ColBERT uses the MaxSim mechanism for scoring:
\begin{equation}
s_{\mathrm{col}}(q,d) = \sum_{i = 1}^{|q|}\ \max_{j \le |d|}\ \cos\!\left(\mathbf{e}_{i}^{q},\mathbf{e}_{j}^{d}\right)
\end{equation}
where $\mathbf{e}_{i}^{q}$ and $\mathbf{e}_{j}^{d}$ denote token embeddings of the query and window \citep{khattab_colbert_2020}. After coarse ranking, BGE-reranker-v2-m3 serves as a cross-encoder for fine-grained ranking on the top-$K$ candidates \citep{chen_bge_m3_2024}. The two-stage pipeline reduces computational cost while balancing precise term matching and semantic understanding.

\paragraph{Evidence-grade-aware Retrieval.} The corpus was partitioned by evidence grade into Grade A and Grades B--E, forming two candidate pools. Candidate generation and truncation were performed separately on each pool. The candidate sets were then merged, and BETR calibration was applied for final global ranking: $r(q,d)=\hat{a}\,s(q,d)+\hat{u}_{\mathrm{Grade}(d)}$. This design enables the system to prioritize higher-grade evidence when candidate relevance is comparable.

\subsection{Benchmark Construction}

To facilitate automated evaluation, we created 1,637 QA pairs. Using the annotated corpus, we performed stratified sampling by evidence grade to obtain candidate evidence windows. GPT-4o \citep{openai_gpt4o_2024} generated a clinical question for each window, strictly grounded in the window's core conclusion, while simultaneously extracting PICOT elements; inclusion of unsupported information was prohibited \citep{saad_falcon_ares_2024}. We then applied round-trip consistency filtering \citep{saad_falcon_ares_2024}: using a system-agnostic hybrid retrieval baseline, questions whose seed window was recalled within top-$K$ entered the main split; the remainder entered a challenge split requiring closer manual inspection.

For reference evidence, we reranked candidates by relevance and had DeepSeek-V3 grade the evidence relationship; strongly supportive and supportive windows formed the gold evidence set. Reference answers were organized by evidence grade, and each answer was decomposed into atomic facts (nuggets) for automated assessment \citep{min_factscore_2023}. Four graduate students reviewed all QA pairs from both splits, discarding questions that were clinically irrelevant or lacked practical value, and retaining 1,637 from approximately 2,000 candidates. To validate that LLM-assisted gold labels are reliable, we additionally had two independent annotators re-label gold evidence and PICOT fields from scratch on an 80-query stratified subset (\S\ref{sec:human-eval}). The benchmark and knowledge graph are available at \url{https://anonymous.4open.science/r/sr-rag-release-4296/}.

\subsection{Evaluation}

SR-RAG evaluation comprised automated and manual dimensions. For automated evaluation, we used a held-out test set ($n=327$), disjoint from BETR training and validation splits, with five metrics: (1) Evidence recall at 10 (R@10): fraction of gold evidence windows retrieved within the top-10 candidates, a retrieval-level metric independent of LLM judges; (2) Nugget coverage (NC): extent to which the answer covers GT core factual units; (3) Answer faithfulness (Faith.): whether statements are supported by retrieved evidence; (4) Semantic similarity (SS): cosine similarity between answer and GT embeddings; (5) PICOT match accuracy (PM, detailed below).

Mainstream RAG evaluation methods currently lack specific metrics for medical QA, making it difficult to detect PICO mismatches. To address this, we developed PICOT match accuracy (PM). Gold PICOT fields were extracted from reference answers by GPT-4o and spot-checked by annotators; only fields unambiguously stated in the source text were retained. For each system output, PICOT fields are extracted and matched field-by-field against gold fields (synonymous expressions permitted; null gold fields excluded). Let $\mathcal{F}_q \subseteq \{P,I,C,O,T\}$ denote the non-null gold fields for query $q$ and $m_f \in \{0,1\}$ the match indicator:
\begin{equation}
\mathrm{PM} = \frac{1}{N}\sum_{q=1}^{N}\frac{\sum_{f \in \mathcal{F}_q} m_f}{|\mathcal{F}_q|}
\end{equation}
i.e., the macro-averaged field match rate across queries.

Nugget coverage and PICOT match accuracy were implemented via LLM-as-judge \citep{zheng_llm_judge_2023}. Answer faithfulness and semantic similarity were assessed via the RAGAS framework \citep{es_ragas_2024}.

For manual evaluation, five sports rehabilitation experts reviewed 20 randomly sampled questions on a five-point Likert scale across five dimensions: medical factual accuracy, answer faithfulness, answer relevance, safety, and PICOT alignment.

\section{Experiments}

\subsection{Knowledge Graph Construction}

Following corpus preprocessing and evidence-grade annotation, we constructed a knowledge graph for sports rehabilitation based on the Youtu-GraphRAG framework. The schema for nodes, relations, and attributes was replaced with PICO-related terms to explicitly encode evidence-based query structure during graph construction (Appendix~C). The final knowledge graph consisted of 21 types of sports rehabilitation disease corpora, together with general cross-disease guidelines, comprising a total of 357,844 nodes and 371,226 edges. Of all nodes, 44,033 (12.3\%) were core medical entities directly aligned with the PICO framework. Among these, \texttt{Intervention} nodes were most prevalent (13,866; 31.5\%), followed by \texttt{Condition} (9,979; 22.7\%), \texttt{Outcome} (8,609; 19.5\%), and \texttt{Population} (6,838; 15.5\%). \texttt{Arm}, \texttt{Device}, and \texttt{Comparator} accounted for 2,134 (4.9\%), 1,569 (3.6\%), and 1,038 (2.4\%), respectively. The graph data will be released upon publication.

\begin{table}[t]
\centering
\footnotesize
\caption{BETR reranking quality on 327 test queries ($K{=}12$).}
\label{tab:betr-rerank}
\setlength{\tabcolsep}{3pt}
\begin{tabular}{@{}lcccc@{}}
\toprule
Method & AvgGrade & HGSR & SA-NDCG & PrefAcc \\
\midrule
Semantic-only & 3.18 & .362 & .781 & .523 \\
Heuristic & 3.34 & .418 & .814 & .586 \\
BETR & \textbf{3.51} & \textbf{.473} & \textbf{.847} & \textbf{.641} \\
\bottomrule
\end{tabular}
\end{table}

\subsection{BETR Calibration Evaluation}

BETR is designed to calibrate ranking so that higher-grade evidence is promoted within the top-$K$ positions, rather than to optimize end-to-end generation quality. We therefore evaluate its reranking behaviour directly. We partitioned the 1,637 benchmark queries into train ($n=983$), validation ($n=327$), and test ($n=327$) splits. BETR parameters were fitted on the train set with hyperparameters selected on the validation set (training details in Appendix~B). For each held-out query we retrieved a fixed candidate pool and reranked it under three settings: (1)~\textbf{Semantic-only}, ranking by semantic score $s(q,d)$ alone; (2)~\textbf{Heuristic}, adding a linearly spaced grade bias $u_t^{h}=-c\cdot\mathrm{rank}(t)$ with $c$ tuned on the validation set; (3)~\textbf{BETR}, using the learned parameters. We report four ranking-quality metrics at $K{=}12$ (the operational evidence budget): average evidence grade of gold windows in top-$K$ (\textbf{AvgGrade}), proportion of gold windows that are also high-grade (A+B) among top-$K$ (\textbf{HGSR}, High-Grade Supportive Ratio), support-aware grade NDCG using gold-evidence labels as relevance (\textbf{SA-NDCG}), and gold-vs-non-gold preference accuracy (\textbf{PrefAcc}).

As shown in Table~\ref{tab:betr-rerank}, BETR consistently outperforms both baselines across all four metrics, consistent with semantic relevance remaining the primary ranking signal while grade biases provide effective calibration among relevant candidates (the learned scale parameter $a=1.035$ and monotonically decreasing biases are detailed in Appendix~B). The gain over Heuristic further demonstrates that data-driven calibration outperforms hand-tuned mappings.

\subsection{Baseline Comparison and Ablation Studies}

We conducted an end-to-end evaluation of SR-RAG on the 327 held-out test queries, reporting five metrics: evidence recall at 10 (R@10), nugget coverage (NC), answer faithfulness (Faith.), semantic similarity (SS), and PICOT match accuracy (PM). We compared SR-RAG against five baselines and performed ablation experiments on the best-performing model, DeepSeek-V3. All results are shown in Table~\ref{tab:main-results}. The benchmark and knowledge graph are available at \url{https://anonymous.4open.science/r/sr-rag-release-4296/}.

\paragraph{Baselines.} We compared SR-RAG with five baselines, all using DeepSeek-V3 as the generator and the same corpus: (1) \textbf{Naive RAG} with BM25 + dense retrieval and RRF fusion; (2) \textbf{Microsoft GraphRAG} \citep{edge_graphrag_2024} with community-based global and local search; (3) \textbf{LightRAG} \citep{guo_lightrag_2024} with dual-level entity--relationship retrieval; (4) \textbf{Youtu-GraphRAG} \citep{dong_youtu_graphrag_2025}, which shares our knowledge graph but uses a generic medical schema without PICO-guided HyDE or BETR; (5) \textbf{Med-R$^2$} \citep{wang_medr2_2025}, an EBM-aligned RAG with a manually defined linear evidence-hierarchy mapping $f_h(x){=}9{-}(e_x{-}1)$, adapted to our corpus.

\begin{table}[t]
\centering
\small
\caption{Automated evaluation results on the SR-RAG benchmark.}
\label{tab:main-results}
\setlength{\tabcolsep}{3.2pt}
\begin{tabular}{@{}lccccc@{}}
\toprule
Method & R@10$\uparrow$ & NC$\uparrow$ & Faith.$\uparrow$ & SS$\uparrow$ & PM$\uparrow$ \\
\midrule
\multicolumn{6}{l}{\textit{Baselines (DeepSeek-V3)}} \\
Naive RAG & 0.643 & 0.718 & 0.769 & 0.841 & 0.582 \\
MS GraphRAG & 0.698 & 0.749 & 0.784 & 0.856 & 0.621 \\
LightRAG & 0.721 & 0.762 & 0.791 & 0.862 & 0.643 \\
Youtu-GraphRAG & \cellcolor{bestgreen}0.741 & \cellcolor{bestgreen}0.773 & 0.798 & \cellcolor{bestgreen}0.868 & 0.659 \\
Med-R$^2$ & 0.724 & 0.758 & \cellcolor{bestgreen}0.803 & 0.861 & \cellcolor{bestgreen}0.678 \\
\midrule
\multicolumn{6}{l}{\textit{SR-RAG (Ours)}} \\
Baichuan-M2 & 0.812 & 0.740 & 0.785 & 0.806 & 0.755 \\
GPT-4o & 0.812 & 0.825 & \cellcolor{bestblue}\textbf{0.842} & 0.862 & 0.762 \\
DeepSeek-V3 & \cellcolor{bestblue}\textbf{0.812} & \cellcolor{bestblue}\textbf{0.830} & 0.819 & 0.882 & \cellcolor{bestblue}\textbf{0.788} \\
\midrule
\multicolumn{6}{l}{\textit{Ablation (DeepSeek-V3)}} \\
w/o HyDE & 0.764 & 0.819 & 0.801 & 0.879 & 0.723 \\
w/o ColBERT & 0.748 & 0.798 & 0.752 & 0.884 & 0.740 \\
w/o PICO schema & 0.753 & 0.774 & 0.834 & \cellcolor{bestblue}\textbf{0.886} & 0.701 \\
w/o BETR & 0.810 & 0.822 & 0.818 & 0.881 & 0.768 \\
\bottomrule
\end{tabular}
\end{table}

\paragraph{Baseline comparison.} SR-RAG (DeepSeek-V3) outperformed all baselines across all five metrics. On the LLM-independent R@10, SR-RAG achieved 0.812 versus 0.643--0.738 for baselines. Youtu-GraphRAG, which shares SR-RAG's knowledge graph, led baselines on R@10, NC, and SS, confirming the value of graph-structured retrieval. Med-R$^2$, the only EBM-aware baseline, led on Faith and PM, confirming that evidence-hierarchy awareness benefits answer quality; however, SR-RAG still outperformed it on PM by $+$16.2\% (0.788 vs.\ 0.678), indicating that learned BETR calibration with PICO-schema-level alignment surpasses manually defined linear mappings.

\paragraph{Model comparison.} DeepSeek-V3 \citep{deepseek_v3_2024} performed best overall, followed by GPT-4o \citep{openai_gpt4o_2024} (highest faithfulness at 0.842). Although Baichuan-M2 \citep{baichuan_m2_2025} is medically specialized, its smaller scale limits the complex multi-step reasoning required by SR-RAG.

\paragraph{Ablation analysis.} We ablated four components on DeepSeek-V3 (see Appendix~C for schema details). HyDE removal caused the largest R@10 drop ($-$0.048), confirming its direct retrieval benefit. Removing the PICO schema had the largest impact on NC ($-$0.056) and PM ($-$0.087) but slightly increased Faith (+0.015), revealing a coverage--alignment tradeoff that underscores the necessity of domain-specific metrics like PM. ColBERT removal caused the largest Faith drop ($-$0.067). BETR removal barely affected R@10 ($-$0.002) but reduced PM by 0.020, consistent with its role as a post-retrieval evidence-hierarchy calibrator rather than a retrieval component.

\subsection{Human Evaluation}
\label{sec:human-eval}

\paragraph{Expert clinician ratings.} We sampled 20 questions and invited five sports rehabilitation experts for manual evaluation on a 1--5 Likert scale across five dimensions (Table~\ref{tab:human-eval}; procedures in Appendix~E).

\begin{table}[t]
\centering
\small
\caption{Expert clinician evaluation results (5 experts, 20 questions, 1--5 Likert scale). Inter-rater SD reflects mean SD across questions per dimension.}
\label{tab:human-eval}
\setlength{\tabcolsep}{4pt}
\begin{tabular}{@{}lccc@{}}
\toprule
\textbf{Dimension} & \textbf{Mean} & \textbf{SD} & \textbf{Inter-rater SD} \\
\midrule
Medical factual accuracy & 4.71 & 0.50 & 0.46 \\
Answer faithfulness & 4.84 & 0.37 & 0.28 \\
Answer relevance & 4.81 & 0.44 & 0.33 \\
Safety & 4.72 & 0.57 & 0.42 \\
PICOT alignment & 4.66 & 0.76 & 0.63 \\
\bottomrule
\end{tabular}
\end{table}

SR-RAG scored 4.66--4.84 across all dimensions. PICOT alignment showed the largest inter-rater SD (0.63), suggesting sensitivity to question format. To calibrate LLM-based judges, we computed Spearman correlations between automated metrics and corresponding expert dimensions on this 20-question subset: NC vs.\ expert completeness $\rho = 0.68$, Faith.\ vs.\ expert accuracy $\rho = 0.74$, PM vs.\ expert PICOT alignment $\rho = 0.71$ (all $p < 0.01$).

\paragraph{Human-verified subset.} To validate that benchmark quality and system rankings are not artifacts of LLM-assisted gold construction, we stratified-sampled 80 queries from the test set and had two independent annotators re-label gold evidence windows (supportive vs.\ not) and gold PICOT fields from scratch, with a third annotator adjudicating disagreements ($\kappa_{\text{evidence}}=0.76$, $\alpha_{\text{PICOT}}=0.71$). Table~\ref{tab:human-verified} reports R@10 and PM recomputed against these human-verified gold labels.

\begin{table}[t]
\centering
\small
\caption{Human-verified subset ($n=80$). R@10$_{\text{hv}}$ and PM$_{\text{hv}}$ are computed against human-annotated gold evidence and PICOT fields.}
\label{tab:human-verified}
\setlength{\tabcolsep}{5pt}
\begin{tabular}{@{}lcccc@{}}
\toprule
 & R@10 & R@10$_{\text{hv}}$ & PM & PM$_{\text{hv}}$ \\
\midrule
Youtu-GraphRAG & 0.741 & 0.718 & 0.659 & 0.631 \\
Med-R$^2$ & 0.724 & 0.703 & 0.678 & 0.649 \\
SR-RAG (DS-V3) & 0.812 & 0.791 & 0.788 & 0.762 \\
\bottomrule
\end{tabular}
\end{table}

System rankings are fully preserved under human-verified gold labels. The absolute scores decrease slightly (R@10: $-$0.021 on average; PM: $-$0.028), consistent with human annotators applying stricter criteria than LLM-assisted construction. SR-RAG maintains a substantial lead over Med-R$^2$ on PM$_{\text{hv}}$ (0.762 vs.\ 0.649, $+$17.4\%). Together with the LLM-independent R@10 and SS metrics, these results confirm that SR-RAG's advantages are not attributable to evaluation-loop bias.

\section{Conclusion}

We present SR-RAG, an EBM-adapted GraphRAG framework for sports rehabilitation. To address the core question of how to adapt the EBM framework to RAG pipelines, we introduce generalizable components organized around two EBM principles: (1) \textbf{Evidence hierarchy}: evidence-grade annotation at the corpus level, BETR for learning grade biases without predefined weights, and dual-track retrieval that recalls guideline evidence separately to prevent dilution; (2) \textbf{PICO alignment}: a PICO-extended graph schema that encodes clinical query structure at the extraction level, and PICO-guided HyDE that bridges the query--evidence semantic gap with soft PICOT constraints. We also construct the first knowledge graph (357,844 nodes, 371,226 edges) and benchmark (1,637 QA pairs) for sports rehabilitation. Automated evaluation, expert clinician review, and a human-verified gold subset all confirm SR-RAG's reliability.

\section*{Limitations}

First, expert evaluation is limited in scale (20 questions, 5 experts). Second, BETR operates at the study-type level without assessing within-study evidence quality (e.g., risk of bias); integrating fine-grained quality dimensions such as GRADE \citep{wang_quicker_2025} into reranking is a promising direction.

\bibliography{refs}

\begin{thebibliography}{50}
\providecommand{\natexlab}[1]{#1}

\bibitem[{Almaadawy et~al.(2024)Almaadawy, Uretsky, Krittanawong, and
  Birnbaum}]{almaadawy_target_2024}
Omar Almaadawy, Barry~F. Uretsky, Chayakrit Krittanawong, and Yochai Birnbaum.
  2024.
\newblock \href {https://doi.org/10.3390/jcm13185562} {Target {Heart} {Rate}
  {Formulas} for {Exercise} {Stress} {Testing}: {What} {Is} the {Evidence}?}
\newblock \emph{Journal of Clinical Medicine}, 13(18):5562.

\bibitem[{{American College of Sports Medicine}(2025)}]{acsm_guidelines_2025}
{American College of Sports Medicine}. 2025.
\newblock \emph{{ACSM}'s {Guidelines} for {Exercise} {Testing} and
  {Prescription}}, 12th edition.
\newblock Wolters Kluwer.

\bibitem[{Asai et~al.(2024)Asai, Wu, Wang, Sil, and
  Hajishirzi}]{asai_selfrag_2024}
Akari Asai, Zeqiu Wu, Yizhong Wang, Avirup Sil, and Hannaneh Hajishirzi. 2024.
\newblock Self-{RAG}: Learning to retrieve, generate, and critique through
  self-reflection.
\newblock \emph{Proceedings of ICLR 2024}.

\bibitem[{Auer et~al.(2024)Auer, Lysak, Nassar et~al.}]{auer_docling_2024}
Christoph Auer, Maksym Lysak, Ahmed Nassar, and 1 others. 2024.
\newblock \href {https://doi.org/10.48550/arXiv.2408.09869} {Docling technical
  report}.
\newblock \emph{arXiv preprint arXiv:2408.09869}.

\bibitem[{{Baichuan AI}(2025)}]{baichuan_m2_2025}
{Baichuan AI}. 2025.
\newblock Baichuan-{M2}: Scaling medical capability with large verifier system.
\newblock \emph{arXiv preprint arXiv:2509.02208}.

\bibitem[{Barbazi et~al.(2025)Barbazi, Shin, Hiremath, and
  Lauff}]{barbazi_exploring_2025}
Neda Barbazi, Ji~Youn Shin, Gurumurthy Hiremath, and Carlye~Anne Lauff. 2025.
\newblock \href {https://doi.org/10.2196/64814} {Exploring health educational
  interventions for children with congenital heart disease: Scoping review}.
\newblock \emph{JMIR Pediatrics and Parenting}, 8:e64814.

\bibitem[{B\'{e}chard and Ayala(2024)}]{bechard_reducing_2024}
Patrice B\'{e}chard and Orlando~Marquez Ayala. 2024.
\newblock \href {https://doi.org/10.18653/v1/2024.naacl-industry.19} {Reducing
  hallucination in structured outputs via {Retrieval}-{Augmented}
  {Generation}}.
\newblock \emph{Proceedings of NAACL-HLT 2024 (Industry Track)}, pages
  228--238.

\bibitem[{Bradley and Terry(1952)}]{bradley_terry_1952}
Ralph~Allan Bradley and Milton~E. Terry. 1952.
\newblock \href {https://doi.org/10.2307/2334029} {Rank analysis of incomplete
  block designs: {I}. the method of paired comparisons}.
\newblock \emph{Biometrika}, 39(3/4):324--345.

\bibitem[{Burges et~al.(2005)Burges, Shaked, Renshaw
  et~al.}]{burges_ranknet_2005}
Chris Burges, Tal Shaked, Erin Renshaw, and 1 others. 2005.
\newblock \href {https://doi.org/10.1145/1102351.1102363} {Learning to rank
  using gradient descent}.
\newblock \emph{Proceedings of ICML 2005}, pages 89--96.

\bibitem[{Cabello et~al.(2024)Cabello, Martin-Turrero, Akujuobi, S{\o}gaard,
  and Bobed}]{cabello_meg_2024}
Laura Cabello, Carmen Martin-Turrero, Uchenna Akujuobi, Anders S{\o}gaard, and
  Carlos Bobed. 2024.
\newblock \href {https://doi.org/10.48550/arXiv.2411.03883} {{MEG}: Medical
  knowledge-augmented large language models for question answering}.
\newblock \emph{arXiv preprint arXiv:2411.03883}.

\bibitem[{Chase(2022)}]{chase_langchain_2022}
Harrison Chase. 2022.
\newblock \href {https://github.com/langchain-ai/langchain} {{LangChain}}.

\bibitem[{Chen et~al.(2024)Chen, Xiao, Zhang, Luo, Lian, and
  Liu}]{chen_bge_m3_2024}
Jianlyu Chen, Shitao Xiao, Peitian Zhang, Kun Luo, Defu Lian, and Zheng Liu.
  2024.
\newblock \href {https://doi.org/10.18653/v1/2024.findings-acl.137}
  {{M3}-embedding: Multi-linguality, multi-functionality, multi-granularity
  text embeddings through self-knowledge distillation}.
\newblock In \emph{Findings of the Association for Computational Linguistics:
  ACL 2024}, pages 2318--2335.

\bibitem[{Cormack et~al.(2009)Cormack, Clarke, and
  Buettcher}]{cormack_rrf_2009}
Gordon~V. Cormack, Charles L.~A. Clarke, and Stefan Buettcher. 2009.
\newblock \href {https://doi.org/10.1145/1571941.1572114} {Reciprocal rank
  fusion outperforms condorcet and individual rank learning methods}.
\newblock \emph{Proceedings of SIGIR 2009}, pages 758--759.

\bibitem[{{DeepSeek-AI}(2024)}]{deepseek_v3_2024}
{DeepSeek-AI}. 2024.
\newblock {DeepSeek-V3} technical report.
\newblock \emph{arXiv preprint arXiv:2412.19437}.

\bibitem[{Dong et~al.(2026)Dong, An, Yu, Zhang, Luo, Huang, Wu, Yin, and
  Sun}]{dong_youtu_graphrag_2025}
Junnan Dong, Siyu An, Yifei Yu, Qian-Wen Zhang, Linhao Luo, Xiao Huang,
  Yunsheng Wu, Di~Yin, and Xing Sun. 2026.
\newblock \href {https://openreview.net/forum?id=yCtgZ2G39E} {Youtu-{GraphRAG}:
  Vertically unified agents for graph retrieval-augmented complex reasoning}.
\newblock In \emph{The Fourteenth International Conference on Learning
  Representations}.
\newblock Poster.

\bibitem[{Duarte et~al.(2024)Duarte, Marques, Gra\c{c}a
  et~al.}]{duarte_lumberchunker_2024}
Andr\'{e}~V. Duarte, Jo\~{a}o Marques, Miguel Gra\c{c}a, and 1 others. 2024.
\newblock \href {https://doi.org/10.18653/v1/2024.findings-emnlp.377}
  {{LumberChunker}: Long-form narrative document segmentation}.
\newblock \emph{Findings of EMNLP 2024}.

\bibitem[{Edge et~al.(2024)Edge, Trinh, Cheng, Bradley, Chao, Mody, Truitt,
  Metropolitansky, Ness, and Larson}]{edge_graphrag_2024}
Darren Edge, Ha~Trinh, Newman Cheng, Joshua Bradley, Alex Chao, Apurva Mody,
  Steven Truitt, Dasha Metropolitansky, Robert~Osazuwa Ness, and Jonathan
  Larson. 2024.
\newblock \href {https://doi.org/10.48550/arXiv.2404.16130} {From local to
  global: A graph {RAG} approach to query-focused summarization}.
\newblock \emph{arXiv preprint arXiv:2404.16130}.

\bibitem[{Es et~al.(2024)Es, James, Espinosa~Anke, and
  Schockaert}]{es_ragas_2024}
Shahul Es, Jithin James, Luis Espinosa~Anke, and Steven Schockaert. 2024.
\newblock \href {https://doi.org/10.18653/v1/2024.eacl-demo.16} {{RAGAs}:
  {Automated} {Evaluation} of {Retrieval} {Augmented} {Generation}}.
\newblock \emph{Proceedings of EACL 2024: System Demonstrations}, pages
  150--158.

\bibitem[{Gao et~al.(2023)Gao, Ma, Lin, and Callan}]{gao_hyde_2023}
Luyu Gao, Xueguang Ma, Jimmy Lin, and Jamie Callan. 2023.
\newblock \href {https://doi.org/10.18653/v1/2023.acl-long.99} {Precise
  zero-shot dense retrieval without relevance labels}.
\newblock \emph{Proceedings of ACL 2023}, pages 1762--1777.

\bibitem[{Gelman et~al.(2013)Gelman, Carlin, Stern, Dunson, Vehtari, and
  Rubin}]{gelman_bda_2013}
Andrew Gelman, John~B. Carlin, Hal~S. Stern, David~B. Dunson, Aki Vehtari, and
  Donald~B. Rubin. 2013.
\newblock \emph{Bayesian Data Analysis}, 3rd edition.
\newblock CRC Press.

\bibitem[{Guo et~al.(2025)Guo, Xia, Yu, Ao, and Huang}]{guo_lightrag_2024}
Zirui Guo, Lianghao Xia, Yanhua Yu, Tu~Ao, and Chao Huang. 2025.
\newblock \href {https://doi.org/10.18653/v1/2025.findings-emnlp.568}
  {{LightRAG}: Simple and fast retrieval-augmented generation}.
\newblock In \emph{Findings of the Association for Computational Linguistics:
  EMNLP 2025}, pages 10746--10761.

\bibitem[{Gupta et~al.(2024)Gupta, Ranjan, and Singh}]{gupta_rag_survey_2024}
Shailja Gupta, Rajesh Ranjan, and Surya~Narayan Singh. 2024.
\newblock \href {https://doi.org/10.48550/arXiv.2410.12837} {A comprehensive
  survey of retrieval-augmented generation ({RAG}): Evolution, current
  landscape and future directions}.
\newblock \emph{arXiv preprint arXiv:2410.12837}.

\bibitem[{Gutierrez et~al.(2024)Gutierrez, Zhu, Huang, Kamoi, Gu, and
  Sun}]{gutierrez_hipporag_2024}
Bernal~Jimenez Gutierrez, Yiheng Zhu, Zhiwei Huang, Ryo Kamoi, Xiaochen Gu, and
  Huan Sun. 2024.
\newblock {HippoRAG}: Neurobiologically inspired long-term memory for large
  language models.
\newblock \emph{Advances in Neural Information Processing Systems}, 37.

\bibitem[{Guyatt et~al.(2008)Guyatt, Oxman, Vist et~al.}]{guyatt_grade_2008}
Gordon~H. Guyatt, Andrew~D. Oxman, Gunn~E. Vist, and 1 others. 2008.
\newblock \href {https://doi.org/10.1136/bmj.39489.470347.AD} {{GRADE}: an
  emerging consensus on rating quality of evidence and strength of
  recommendations}.
\newblock \emph{BMJ}, 336(7650):924--926.

\bibitem[{Hager et~al.(2024)Hager, Jungmann, Holland
  et~al.}]{hager_evaluation_2024}
Paul Hager, Friederike Jungmann, Robbie Holland, and 1 others. 2024.
\newblock \href {https://doi.org/10.1038/s41591-024-03097-1} {Evaluation and
  mitigation of the limitations of large language models in clinical
  decision-making}.
\newblock \emph{Nature Medicine}, 30(9):2613--2622.

\bibitem[{{Interamerican Society of Cardiology
  (SIAC)}(2024)}]{noauthor_2024_2024}
{Interamerican Society of Cardiology (SIAC)}. 2024.
\newblock \href {https://doi.org/10.1016/j.rec.2024.05.001} {2024 {SIAC}
  guidelines on cardiorespiratory rehabilitation in pediatric patients with
  congenital heart disease}.
\newblock \emph{Revista Espa\~{n}ola de Cardiolog\'{i}a (English Edition)},
  77(8):680--689.

\bibitem[{Jain et~al.(2025)Jain, Aggarwal, and Saladi}]{jain_autochunker_2025}
Arihant Jain, Purav Aggarwal, and Anoop Saladi. 2025.
\newblock \href {https://doi.org/10.18653/v1/2025.acl-industry.69}
  {{AutoChunker}: Structured text chunking and its evaluation}.
\newblock \emph{Proceedings of ACL 2025 (Industry Track)}, pages 983--995.

\bibitem[{{Joanna Briggs Institute}(2014)}]{jbi_levels_2014}
{Joanna Briggs Institute}. 2014.
\newblock {JBI} levels of evidence.

\bibitem[{Khattab and Zaharia(2020)}]{khattab_colbert_2020}
Omar Khattab and Matei Zaharia. 2020.
\newblock \href {https://doi.org/10.1145/3397271.3401075} {{ColBERT}: Efficient
  and effective passage search via contextualized late interaction over
  {BERT}}.
\newblock \emph{Proceedings of SIGIR 2020}, pages 39--48.

\bibitem[{Lauer et~al.(1999)Lauer, Francis, Okin, Pashkow, Snader, and
  Marwick}]{lauer_impaired_1999}
Michael~S. Lauer, Gary~S. Francis, Peter~M. Okin, Fredric~J. Pashkow, Claire~E.
  Snader, and Thomas~H. Marwick. 1999.
\newblock \href {https://doi.org/10.1001/jama.281.6.524} {Impaired
  {Chronotropic} {Response} to {Exercise} {Stress} {Testing} as a {Predictor}
  of {Mortality}}.
\newblock \emph{JAMA}, 281(6):524--529.

\bibitem[{Lewis et~al.(2020)Lewis, Perez, Piktus, Petroni, Karpukhin, Goyal,
  K\"{u}ttler, Lewis, Yih, Rockt\"{a}schel, Riedel, and
  Kiela}]{lewis_retrieval-augmented_2020}
Patrick Lewis, Ethan Perez, Aleksandra Piktus, Fabio Petroni, Vladimir
  Karpukhin, Naman Goyal, Heinrich K\"{u}ttler, Mike Lewis, Wen-tau Yih, Tim
  Rockt\"{a}schel, Sebastian Riedel, and Douwe Kiela. 2020.
\newblock Retrieval-{Augmented} {Generation} for {Knowledge}-{Intensive} {NLP}
  {Tasks}.
\newblock \emph{Advances in Neural Information Processing Systems},
  33:9459--9474.

\bibitem[{Lu et~al.(2025)Lu, Liang, Pan, Zhang, Dong, Wu, Leng, Cui, and
  Zhang}]{wang_medr2_2025}
Keer Lu, Zheng Liang, Da~Pan, Shusen Zhang, Guosheng Dong, Zhonghai Wu, Huang
  Leng, Bin Cui, and Wentao Zhang. 2025.
\newblock \href {https://doi.org/10.48550/arXiv.2501.11885} {{Med-R$^2$}:
  Crafting trustworthy {LLM} physicians via retrieval and reasoning of
  evidence-based medicine}.
\newblock \emph{arXiv preprint arXiv:2501.11885}.
\newblock Accepted to The Web Conference 2026 (Research Track).

\bibitem[{Min et~al.(2023)Min, Krishna, Lyu et~al.}]{min_factscore_2023}
Sewon Min, Kalpesh Krishna, Xinxi Lyu, and 1 others. 2023.
\newblock \href {https://doi.org/10.18653/v1/2023.emnlp-main.741} {{FActScore}:
  Fine-grained atomic evaluation of factual precision in long form text
  generation}.
\newblock \emph{Proceedings of EMNLP 2023}, pages 12076--12100.

\bibitem[{{National Health and Medical Research
  Council}(2009)}]{nhmrc_evidence_2009}
{National Health and Medical Research Council}. 2009.
\newblock {NHMRC} additional levels of evidence and grades for recommendations.
\newblock Technical report, Australian Government.

\bibitem[{{OCEBM Levels of Evidence Working Group}(2011)}]{ocebm_levels_2011}
{OCEBM Levels of Evidence Working Group}. 2011.
\newblock \href
  {https://www.cebm.ox.ac.uk/resources/levels-of-evidence/ocebm-levels-of-evidence}
  {The {Oxford} {Levels} of {Evidence} 2}.
\newblock Oxford Centre for Evidence-Based Medicine.

\bibitem[{{OpenAI}(2024)}]{openai_gpt4o_2024}
{OpenAI}. 2024.
\newblock {GPT-4o} system card.
\newblock \emph{arXiv preprint arXiv:2410.21276}.

\bibitem[{Saad-Falcon et~al.(2024)Saad-Falcon, Khattab, Potts, and
  Zaharia}]{saad_falcon_ares_2024}
Jon Saad-Falcon, Omar Khattab, Christopher Potts, and Matei Zaharia. 2024.
\newblock \href {https://doi.org/10.18653/v1/2024.naacl-long.20} {{ARES}: An
  automated evaluation framework for retrieval-augmented generation systems}.
\newblock In \emph{Proceedings of the 2024 Conference of the North American
  Chapter of the Association for Computational Linguistics: Human Language
  Technologies (Volume 1: Long Papers)}, pages 338--354.

\bibitem[{Sackett et~al.(1996)Sackett, Rosenberg, Gray, Haynes, and
  Richardson}]{sackett_ebm_1996}
David~L. Sackett, William M.~C. Rosenberg, J.~A.~Muir Gray, R.~Brian Haynes,
  and W.~Scott Richardson. 1996.
\newblock \href {https://doi.org/10.1136/bmj.312.7023.71} {Evidence based
  medicine: what it is and what it isn't}.
\newblock \emph{BMJ}, 312(7023):71--72.

\bibitem[{Sun et~al.(2025{\natexlab{a}})Sun, Zhao, Chen, and
  Qin}]{sun_picos_rag_2025}
Mengzhou Sun, Sendong Zhao, Jianyu Chen, and Bin Qin. 2025{\natexlab{a}}.
\newblock \href {https://doi.org/10.48550/arXiv.2510.23998} {{PICOs-RAG}:
  {PICO}-supported query rewriting for retrieval-augmented generation in
  evidence-based medicine}.
\newblock \emph{arXiv preprint arXiv:2510.23998}.

\bibitem[{Sun et~al.(2025{\natexlab{b}})Sun, Zhao, Chen, Wang, and
  Qin}]{sun_metarag_2025}
Mengzhou Sun, Sendong Zhao, Jianyu Chen, Haochun Wang, and Bing Qin.
  2025{\natexlab{b}}.
\newblock \href {https://doi.org/10.48550/arXiv.2510.24003} {{META-RAG}:
  Meta-analysis-inspired evidence-re-ranking method for retrieval-augmented
  generation in evidence-based medicine}.
\newblock \emph{arXiv preprint arXiv:2510.24003}.

\bibitem[{Takehi et~al.(2025)Takehi, Clavi\'{e}, Lee, and
  Shakir}]{takehi_mxbai_edge_2025}
Rikiya Takehi, Benjamin Clavi\'{e}, Sean Lee, and Aamir Shakir. 2025.
\newblock \href {https://doi.org/10.48550/arXiv.2510.14880} {Fantastic (small)
  retrievers and how to train them: mxbai-edge-colbert-v0 tech report}.
\newblock \emph{arXiv preprint arXiv:2510.14880}.

\bibitem[{Ubeda~Tikkanen et~al.(2023)Ubeda~Tikkanen, Vova, Holman
  et~al.}]{ubeda_tikkanen_core_2023}
Ana Ubeda~Tikkanen, Joshua Vova, Lainie Holman, and 1 others. 2023.
\newblock \href {https://doi.org/10.3389/fped.2023.1104794} {Core components of
  a rehabilitation program in pediatric cardiac disease}.
\newblock \emph{Frontiers in Pediatrics}, 11.

\bibitem[{Wang et~al.(2023)Wang, Yang, and Wei}]{wang_query2doc_2023}
Liang Wang, Nan Yang, and Furu Wei. 2023.
\newblock \href {https://doi.org/10.18653/v1/2023.emnlp-main.585} {Query2doc:
  Query expansion with large language models}.
\newblock \emph{Proceedings of EMNLP 2023}, pages 9414--9423.

\bibitem[{Wang et~al.(2025)Wang, Yang, Liu et~al.}]{wang_quicker_2025}
Xiangyu Wang, Kaiping Yang, Xucheng Liu, and 1 others. 2025.
\newblock \href {https://doi.org/10.1038/s41746-025-02273-y} {Streamlining
  evidence based clinical recommendations with large language models}.
\newblock \emph{npj Digital Medicine}, 8(1):793.

\bibitem[{Xiong et~al.(2024)Xiong, Jin, Lu, and Zhang}]{xiong_mirage_2024}
Guangzhi Xiong, Qiao Jin, Zhiyong Lu, and Aidong Zhang. 2024.
\newblock \href {https://doi.org/10.18653/v1/2024.findings-acl.372}
  {Benchmarking retrieval-augmented generation for medicine}.
\newblock \emph{Findings of ACL 2024}, pages 6233--6251.

\bibitem[{Yan et~al.(2024)Yan, Gu, Zhu, and Ling}]{yan_crag_2024}
Shi-Qi Yan, Jia-Chen Gu, Yun Zhu, and Zhen-Hua Ling. 2024.
\newblock Corrective retrieval augmented generation.
\newblock In \emph{Proceedings of the Twelfth International Conference on
  Learning Representations (ICLR 2024)}.

\bibitem[{Yang et~al.(2025)Yang, Ning, Keppo et~al.}]{yang_rag_healthcare_2025}
Rui Yang, Yilin Ning, Emmi Keppo, and 1 others. 2025.
\newblock \href {https://doi.org/10.1038/s44401-024-00004-1}
  {Retrieval-augmented generation for generative artificial intelligence in
  health care}.
\newblock \emph{npj Health Systems}, 2:2.

\bibitem[{Zhao et~al.(2024)Zhao, Ji, Qi et~al.}]{zhao_metachunking_2024}
Jihao Zhao, Zhiyuan Ji, Pengnian Qi, and 1 others. 2024.
\newblock Meta-chunking: Learning text segmentation and semantic completion via
  logical perception.
\newblock \emph{arXiv preprint arXiv:2410.12788}.

\bibitem[{Zheng et~al.(2023)Zheng, Chiang, Sheng et~al.}]{zheng_llm_judge_2023}
Lianmin Zheng, Wei-Lin Chiang, Ying Sheng, and 1 others. 2023.
\newblock Judging {LLM}-as-a-judge with {MT-Bench} and chatbot arena.
\newblock \emph{Advances in Neural Information Processing Systems}, 36.

\bibitem[{Zhu et~al.(2024)Zhu, Zhao, Yu, Huang, and Liu}]{zhu_hyqe_2024}
Weichao Zhu, Liang Zhao, Cao Yu, Songlin Huang, and Zhiguo Liu. 2024.
\newblock \href {https://doi.org/10.18653/v1/2024.findings-emnlp.761} {{HyQE}:
  Ranking contexts with hypothetical query embeddings}.
\newblock \emph{Findings of EMNLP 2024}, pages 13024--13036.

\end{thebibliography}

\appendix
\renewcommand{\thefigure}{\Alph{section}.\arabic{figure}}
\renewcommand{\thetable}{\Alph{section}.\arabic{table}}
\renewcommand{\theequation}{\Alph{section}.\arabic{equation}}
\renewcommand{\thealgorithm}{\Alph{section}.\arabic{algorithm}}
\renewcommand{\theHfigure}{\Alph{section}.\arabic{figure}}
\renewcommand{\theHtable}{\Alph{section}.\arabic{table}}
\renewcommand{\theHequation}{\Alph{section}.\arabic{equation}}
\renewcommand{\theHalgorithm}{\Alph{section}.\arabic{algorithm}}

\section{Chunking Strategy}
\setcounter{figure}{0}\setcounter{table}{0}\setcounter{equation}{0}\setcounter{algorithm}{0}
\label{app:chunking}

In the chunking phase, we adopted an LLM-aware hybrid chunking strategy. First, MarkdownHeaderTextSplitter divided documents into sections by heading level \citep{chase_langchain_2022}. Each section was further split into numbered atomic blocks preserving paragraph structure. Under prompt constraints, the LLM performed semantic grouping of atomic blocks and returned group IDs. The program concatenated each group into final evidence windows. This strategy aimed to enhance the semantic integrity and clinical information density of the windows while preserving paragraph boundaries, aligning with recent LLM-aware chunking methods such as AutoChunker, MetaChunking, and LumberChunker \citep{jain_autochunker_2025,zhao_metachunking_2024,duarte_lumberchunker_2024}.

\section{BETR Full Derivation}
\setcounter{figure}{0}\setcounter{table}{0}\setcounter{equation}{0}\setcounter{algorithm}{0}
\label{app:betr}

\subsection{Training Labels}

For each question $q$, candidate windows included in the reference evidence chain serve as positive examples: $\mathcal{C}_q^+=\mathcal{C}_q\cap\mathcal{W}^\star(q)$; the remainder serve as negatives: $\mathcal{C}_q^-=\mathcal{C}_q\setminus\mathcal{W}^\star(q)$. This design avoids per-item manual annotation and aligns the ranking objective with evidence window selection.

\subsection{Pairwise Ranking Objective}

BETR adopts a pairwise learning-to-rank approach \citep{burges_ranknet_2005}, learning preferences through pairwise comparisons. Specifically, for the candidate pool of question $q$, we construct positive--negative pairs $(d^+, d^-)$ with $d^+ \in \mathcal{C}_q\cap\mathcal{W}^\star(q)$ and $d^- \in \mathcal{C}_q\setminus\mathcal{W}^\star(q)$, and model preference probability via the Bradley--Terry model \citep{bradley_terry_1952}:
\begin{equation}
P(d^+ \succ d^- \mid q)=\sigma\big(a\Delta s + u_{t^+}-u_{t^-}\big),
\end{equation}
where $\Delta s = s(q,d^+) - s(q,d^-)$ is the semantic relevance difference; $t^+=\mathrm{Grade}(d^+)$ and $t^-=\mathrm{Grade}(d^-)$ denote evidence grades; $u_t$ is the grade bias; $a > 0$ is the scale parameter; and $\sigma(\cdot)$ is the sigmoid function. This formulation jointly considers two signals: (1) semantic relevance difference $\Delta s$, and (2) evidence grade difference $u_{t^+} - u_{t^-}$. When semantic scores are comparable, windows with higher evidence grades receive additional positive bias.

\subsection{Ordered Hierarchical Parameterization}

To explicitly encode a pyramid-shaped evidence hierarchy and prevent grade inversion under noisy labels, we fix the grade ordering as $A\succ B\succ C\succ D\succ E$ and adopt monotonically constrained incremental parameterization:
\begin{equation}
\begin{aligned}
u_A &= 0,\\
u_B &= -\delta_B,\\
u_C &= -(\delta_B+\delta_C),\\
u_D &= -(\delta_B+\delta_C+\delta_D),\\
u_E &= -(\delta_B+\delta_C+\delta_D+\delta_E).
\end{aligned}
\end{equation}
where $\delta_B,\delta_C,\delta_D,\delta_E\ge 0$. This form naturally guarantees $u_A\ge u_B\ge u_C\ge u_D\ge u_E$, yielding an evidence-grade pyramid consistent with EBM.

We then cast BETR parameter estimation in a Bayesian framework, jointly learning the scale parameter $a$ and grade increments $\delta = (\delta_B, \delta_C, \delta_D, \delta_E)$ via MAP estimation.

\subsection{Prior Distribution}

We impose zero-centered priors on the parameters, encoding the default assumption of no grade bias:
\begin{gather}
\alpha \sim \mathcal{N}(0, \sigma_a^2),\quad
\delta_i \sim \mathcal{N}^+(0, \tau^2),\notag\\
i \in \{B,C,D,E\}
\end{gather}
where $\mathcal{N}^+$ denotes a Gaussian truncated to nonnegative values, ensuring grade monotonicity. This prior has maximum density at $\delta=0$, implying that ranking defaults to being driven by semantic relevance $s(q,d)$. Meanwhile, the prior for $\alpha$ is centered at $0$, corresponding to the default $a=\exp(\alpha)\approx 1$, meaning that semantic scores and grade biases are summed on the same scale. This prior assumption aligns with standard reranking, which orders results solely by relevance scores.

\subsection{Likelihood Function}

Given parameters $(\alpha, \delta)$, the likelihood of the observed pairwise preference data ($d^+ \succ d^-$) is
\begin{multline}
P(\mathcal{D} \mid \alpha, \boldsymbol{\delta})
= \prod_{q \in \mathcal{Q}_{\mathrm{train}}}\;
  \prod_{(d^+,d^-) \in \mathcal{P}_q}\\
  \sigma\!\bigl(a\,\Delta s + u_{t^+} - u_{t^-}\bigr),
\end{multline}
where $a = \exp(\alpha)$ and $\Delta s = s(q, d^+) - s(q, d^-)$.

\subsection{Posterior and MAP Estimation}

According to Bayes' theorem \citep{gelman_bda_2013}, the posterior distribution is proportional to the product of the likelihood and priors:
\begin{equation}
P(\alpha, \delta \mid \mathcal{D}) \propto P(\mathcal{D} \mid \alpha, \delta) \cdot P(\alpha) \cdot P(\delta).
\end{equation}
Taking the logarithm and normalizing, the MAP objective is equivalent to
\begin{equation}
\begin{split}
\max_{\alpha,\;\boldsymbol{\delta}\ge\mathbf{0}}\;&
\frac{1}{|\mathcal{Q}_{\mathrm{tr}}|}
\!\sum_{q\in\mathcal{Q}_{\mathrm{tr}}}
\frac{1}{\max(1,|\mathcal{P}_q|)}\\
&\times\!\sum_{(d^+,d^-)\in\mathcal{P}_q}
\!\log\sigma\!\bigl(a\,\Delta s + u_{t^+}\!-\!u_{t^-}\bigr)\\
&-\frac{1}{2\tau^2}\bigl(\delta_B^2\!+\!\delta_C^2\!+\!\delta_D^2\!+\!\delta_E^2\bigr)
-\frac{\alpha^2}{2\sigma_a^2}.
\end{split}
\end{equation}
The first term is the query-normalized pairwise log-likelihood, where we average within each query to avoid overweighting queries with more sampled pairs. The remaining terms are quadratic shrinkage penalties corresponding to the Gaussian priors on $\alpha$ and $\delta$, respectively.

\subsection{Framework Advantages}

This framework offers several advantages: (1) the prior provides interpretable default behavior, treating grade biases as minimum necessary adjustments; (2) the posterior is likelihood-dominated when data are ample and shrinks toward the prior when data are scarce, achieving adaptive regularization; (3) the hyperparameter $\tau$ is selected via grid search on a validation set and fixed across all experiments, offering high interpretability: it quantifies prior confidence in the assumed grade effects, thereby minimizing manual intervention.

\subsection{Training Details}

We split the 1,637 benchmark queries into three non-overlapping sets: $\mathcal{Q}_{\mathrm{train}}$ ($n=983$) for fitting calibrator parameters, $\mathcal{Q}_{\mathrm{val}}$ ($n=327$) for hyperparameter selection and early stopping via grid search, and $\mathcal{Q}_{\mathrm{test}}$ ($n=327$) reserved exclusively for end-to-end evaluation. Splitting was query-grouped, ensuring that candidate windows for the same query did not span sets, thereby preventing information leakage. During BETR training, candidate sets $\mathcal{C}_q$ were constructed using the same candidate generation process $\mathrm{Cand}(\cdot)$ as at inference, reducing distribution shift.

\subsection{BETR Training Curves and Learned Parameters}

\begin{figure}[t]
\centering
\includegraphics[width=\columnwidth]{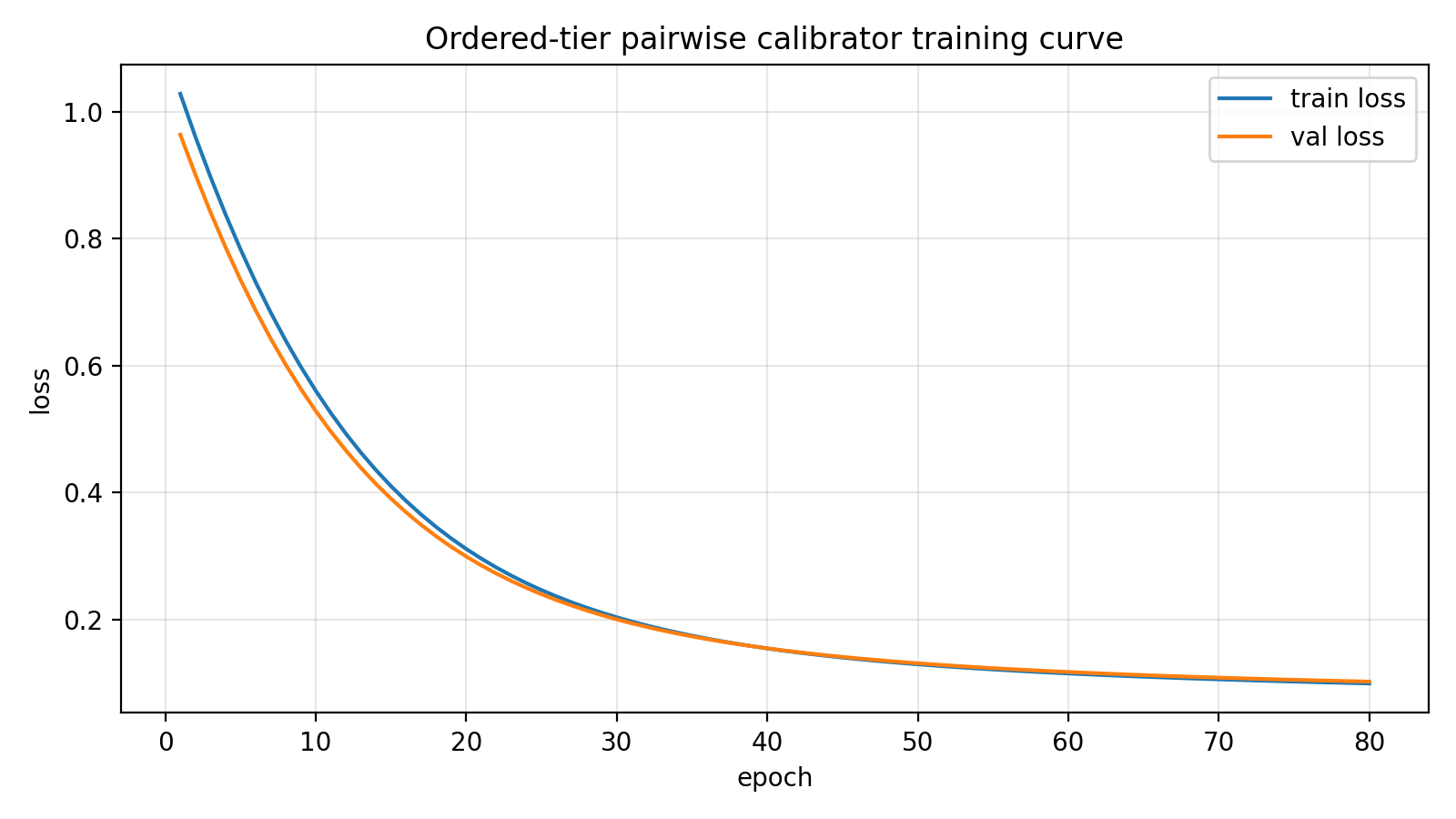}
\caption{BETR training curves showing MAP objective convergence and learned grade bias parameters.}
\label{fig:betr-training}
\end{figure}

Fig.~\ref{fig:betr-training} shows the training dynamics of BETR. The learned parameters are summarized in Table~\ref{tab:betr-params}.

\begin{table}[t]
\centering
\small
\caption{Learned BETR parameters.}
\label{tab:betr-params}
\begin{tabular}{lcc}
\toprule
\textbf{Parameter} & \textbf{Symbol} & \textbf{Value} \\
\midrule
Scale & $a = \exp(\alpha)$ & 1.0348 \\
Grade A bias & $u_A$ & 0 (anchor) \\
Grade B bias & $u_B$ & $-0.1287$ \\
Grade C bias & $u_C$ & $-0.2575$ \\
Grade D bias & $u_D$ & $-0.3863$ \\
Grade E bias & $u_E$ & $-0.5151$ \\
Shrinkage $\tau$ & selected & 1.0 \\
Scale prior $\sigma_a$ & fixed & 5.0 \\
\bottomrule
\end{tabular}
\end{table}

The near-unity scale parameter ($a \approx 1.035$) confirms that BETR preserves the original semantic relevance scale. The grade bias decrements ($\approx 0.129$ per tier) form an approximately uniform staircase, consistent with the pyramid-shaped evidence hierarchy assumed in EBM.

\section{Schema Specifications and Community Compression}
\setcounter{figure}{0}\setcounter{table}{0}\setcounter{equation}{0}\setcounter{algorithm}{0}
\label{app:schema}

\subsection{Schema Self-expansion}

Youtu-GraphRAG allows agents to propose schema extensions based on document content:
\begin{gather}
\Delta \mathcal{S}
=\bigl\langle \Delta \mathcal{S}_e,\;\Delta \mathcal{S}_r,\;
\Delta \mathcal{S}_{\text{attr}}\bigr\rangle\notag\\
=\mathbb{I}\!\bigl[f_{\mathrm{LLM}}(x,\mathcal{S})\odot\mathcal{S}\bigr]\ge\mu,
\end{gather}
where $\mu$ is the confidence threshold for accepting new schema elements, and $\Delta \mathcal{S}$ contains candidate extensions for entities, relations, and attributes. For our task, this mechanism enables graph and index updates across disease subtypes.

\subsection{Dually-perceived Community Compression and Knowledge Tree Indexing}

To reduce density and noise in the raw triple graph and shorten retrieval context, we adopted Youtu-GraphRAG's dually-perceived community detection. The affinity between node $e_i$ and community $\mathcal{C}_m$ is defined as
\begin{equation}
\phi(e_i,\mathcal{C}_m)=\underbrace{\mathbb{S}_r(e_i,\mathcal{C}_m)}_{\text{relational}}\ \oplus\ \lambda\,\underbrace{\mathbb{S}_s(e_i,\mathcal{C}_m)}_{\text{semantic}},
\end{equation}
where $\mathbb{S}_r$ measures Jaccard similarity over relation-type sets $\Psi(\cdot)$, $\mathbb{S}_s$ measures subgraph semantic similarity, and $\lambda$ is a trade-off coefficient. Iteratively, the algorithm performs pairwise community merging based on the threshold criterion:
\begin{equation}
\mathbb{E}\!\left[\phi\!\left(e_i,\mathcal{C}_a^{(t)}\right)\right]-\mathbb{E}\!\left[\phi\!\left(e_i,\mathcal{C}_b^{(t)}\right)\right] < \epsilon.
\end{equation}
Ultimately, the community structure forms a four-layer knowledge tree $\mathcal{K}=\bigcup_{\ell=1}^{4} L_\ell$: $L_4$ = community, $L_3$ = keyword, $L_2$ = entity--relation triple, $L_1$ = attribute.

\subsection{Schema Tables}

Table~\ref{tab:app-pico-schema} lists the PICO-extended schema used for graph construction and retrieval. Table~\ref{tab:app-ablation-schema} lists the PICO-neutral schema used in the \emph{w/o PICO-extended schema} ablation.

\begin{table*}[t]
\centering
\small
\caption{PICO-extended schema used in graph construction and retrieval.}
\label{tab:app-pico-schema}
\begin{tabular}{p{0.22\textwidth}p{0.34\textwidth}p{0.36\textwidth}}
\toprule
\textbf{Nodes} & \textbf{Relations} & \textbf{Attributes} \\
\midrule
\texttt{Population}, \texttt{Condition}, \texttt{Intervention}, \texttt{Comparator}, \texttt{Outcome}, \texttt{Timepoint}, \texttt{Arm}, \texttt{Device}, \texttt{Recommendation}, \texttt{AdverseEvent}, \texttt{Contraindication}
&
\texttt{has\_population}, \texttt{has\_condition}, \texttt{uses\_intervention}, \texttt{compares\_to}, \texttt{reports\_outcome}, \texttt{has\_timepoint}, \texttt{targets\_arm}, \texttt{uses\_device}, \texttt{recommends\_for}, \texttt{recommends\_against}, \texttt{has\_adverse\_event}, \texttt{contraindicated\_for}
&
\texttt{age\_bin}, \texttt{age\_range}, \texttt{sex}, \texttt{baseline\_severity}, \texttt{time\_since\_injury\_or\_surgery}, \texttt{inclusion\_criteria}, \texttt{exclusion\_criteria}, \texttt{dose}, \texttt{frequency}, \texttt{session\_duration}, \texttt{intensity}, \texttt{progression\_rule}, \texttt{setting}, \texttt{supervision}, \texttt{followup\_weeks}, \texttt{timepoint\_value}, \texttt{timepoint\_unit}, \texttt{measure\_name}, \texttt{outcome\_domain}, \texttt{primary\_outcome}, \texttt{adverse\_event}, \texttt{contraindication}, \texttt{recommendation\_strength}, \texttt{evidence\_certainty}, \texttt{applicability\_notes}, \texttt{study\_design}, \texttt{sample\_size}, \texttt{protocol\_params}
\\
\bottomrule
\end{tabular}
\end{table*}

\begin{table*}[t]
\centering
\small
\caption{PICO-neutral schema used in the \emph{w/o PICO-extended schema} ablation.}
\label{tab:app-ablation-schema}
\begin{tabular}{p{0.22\textwidth}p{0.34\textwidth}p{0.36\textwidth}}
\toprule
\textbf{Nodes} & \textbf{Relations} & \textbf{Attributes} \\
\midrule
\texttt{ClinicalConcept}, \texttt{EvidenceStatement}, \texttt{Study}, \texttt{Guideline}, \texttt{Recommendation}, \texttt{Protocol}, \texttt{OutcomeMeasure}, \texttt{Device}, \texttt{AdverseEvent}, \texttt{Contraindication}, \texttt{Topic}
&
\texttt{mentions}, \texttt{associated\_with}, \texttt{belongs\_to\_topic}, \texttt{reported\_in}, \texttt{supported\_by}, \texttt{has\_protocol}, \texttt{uses\_device}, \texttt{reports\_measure}, \texttt{recommends\_for}, \texttt{recommends\_against}, \texttt{has\_adverse\_event}, \texttt{contraindicated\_for}
&
\texttt{participant\_characteristics}, \texttt{procedure\_details}, \texttt{measurement\_details}, \texttt{temporal\_details}, \texttt{adverse\_event}, \texttt{contraindication}, \texttt{recommendation\_strength}, \texttt{evidence\_certainty}, \texttt{applicability\_notes}, \texttt{study\_design}, \texttt{sample\_size}, \texttt{protocol\_params}
\\
\bottomrule
\end{tabular}

\vspace{0.5em}
\noindent\small\textit{Note.} The first four attributes are free-text fields that collapse the 20 PICO-specific attributes of Table~\ref{tab:app-pico-schema} into domain-generic buckets (\texttt{participant\_characteristics} $\leftarrow$ P-related; \texttt{procedure\_details} $\leftarrow$ I-related; \texttt{measurement\_details} $\leftarrow$ O-related; \texttt{temporal\_details} $\leftarrow$ T-related), preserving underlying clinical information while removing PICO-structured encoding.
\end{table*}

\section{Pipeline Details and Benchmark Construction}
\setcounter{figure}{0}\setcounter{table}{0}\setcounter{equation}{0}\setcounter{algorithm}{0}
\label{app:pipeline}

\subsection{Three-channel Retrieval Fusion}

SR-RAG combines ranking results from three retrieval channels using Reciprocal Rank Fusion (RRF) \citep{cormack_rrf_2009}. For any candidate window $d$, the RRF score is defined as:
\begin{equation}
\mathrm{RRF}(d)=\sum_{c\in\{\mathrm{Dense},\mathrm{Graph},\mathrm{HyDE}\}} \frac{1}{k+\mathrm{rank}_c(d)},
\end{equation}
where $\mathrm{rank}_c(d)$ is the rank of window $d$ in channel $c$, and $k$ is a smoothing constant. After fusion, the pipeline follows Youtu-GraphRAG's parallel retrieval strategy (Entity Matching, Triple Matching, Community Filtering) to generate candidate windows.

\subsection{HyDE Background}

The original HyDE method addresses poor zero-shot dense retrieval performance without relevance annotations \citep{gao_hyde_2023}. LLMs generate several hypothetical documents related to a user query, which are subsequently used for dense retrieval in the corpus. This paradigm tolerates factual errors in hypothetical documents: because retrieval relies on dense embeddings rather than lexical matching, semantic similarity is preserved even when generated content contains inaccuracies \citep{gao_hyde_2023}. Subsequent work identified another advantage of HyDE: bridging the semantic gap between queries and documents \citep{wang_query2doc_2023}. In RAG, document-form inputs generally outperform raw question-form queries; thus, HyDE effectively improves retrieval quality and downstream performance.

\subsection{ColBERT MaxSim Mechanism}

ColBERT uses the MaxSim mechanism for scoring:
\begin{equation}
s_{\mathrm{col}}(q,d) = \sum_{i = 1}^{|q|}\ \max_{j \le |d|}\ \cos\!\left(\mathbf{e}_{i}^{q},\mathbf{e}_{j}^{d}\right)
\end{equation}
Here, $\mathbf{e}_{i}^{q}$ and $\mathbf{e}_{j}^{d}$ denote the $i$-th and $j$-th token embeddings of the query and window, respectively. The MaxSim mechanism computes, for each query token, its maximum similarity with any window token and sums across tokens \citep{khattab_colbert_2020}, improving sensitivity to medical abbreviations, scale names, and synonyms while increasing matching robustness.

\subsection{Cross-encoder Details}

After coarse ranking, we used BGE-reranker-v2-m3 as a cross-encoder to perform fine-grained ranking on the top-$K$ high-scoring candidates \citep{chen_bge_m3_2024}. The cross-encoder jointly encodes the query and each candidate window via concatenation, enabling full interaction through multiple self-attention layers and outputting an overall relevance logit $s(q,d)$. The computational cost of the cross-encoder increases linearly with the number of candidates. By adopting a two-stage pipeline, the system reduces reranking time and computational cost while balancing precise term matching and semantic understanding of long texts.

\subsection{Benchmark Construction Full Pipeline}

\paragraph{Stage 1: Question Generation.} Using the corpus with completed evidence-grade annotation and chunking, we performed stratified sampling by evidence grade to obtain candidate evidence windows, filtering those with insufficient information density or lacking substantive clinical conclusions. For each candidate window, GPT-4o \citep{openai_gpt4o_2024} generated a clinical question strictly corresponding to the window's core conclusion while simultaneously extracting PICOT elements. During generation, questions had to be directly supported by the window; inclusion of information not present in the window was prohibited \citep{saad_falcon_ares_2024}. Additionally, the model annotated evidence certainty (sufficient or uncertain) for each question, enabling stratified evaluation and error analysis.

\paragraph{Stage 2: Retrieval Accessibility Check.} We performed retrieval accessibility checks for each generated question, similar to the round-trip consistency filtering in ARES \citep{saad_falcon_ares_2024}. Using a system-agnostic hybrid retrieval baseline (combining sparse and dense methods), we retrieved the top-$K$ candidate windows. If the seed window was recalled within top-$K$, the question was deemed ``accessible'' and entered the main split; otherwise, it entered the challenge split for manual review.

\paragraph{Stage 3: Reference Evidence Construction.} We selected gold windows for each accessible question from a larger candidate pool. This entailed reranking candidates by relevance, then having DeepSeek-V3 \citep{deepseek_v3_2024} grade the evidence relationship (strongly supportive/supportive/weakly related/unrelated). Strongly supportive and supportive windows formed the gold evidence set. Based on gold windows, we generated hierarchical reference answers: an exact answer summarizing the core conclusion, and an ideal answer organized by evidence grades (A--E), with Grade A prioritized and remaining evidence presented in descending pyramid order. All statements were strictly aligned with gold windows, with source identifiers for traceability. GPT-4o then decomposed the exact answer into atomic facts (nuggets) for automated assessment \citep{min_factscore_2023}.

\paragraph{Stage 4: Human Review.} Four sports science graduate students manually reviewed all QA pairs from both the main and challenge splits, with particular attention to samples marked ``evidence uncertain'' and those in the challenge split. Questions that were clinically irrelevant, lacked practical rehabilitation value, or contained factual errors were discarded. Approximately 2,000 questions were generated initially; after screening, 1,637 were retained as the official evaluation set. We extracted questions, exact answers, and nuggets and compiled them into a public dataset for reusability. Because some gold windows originate from institutionally subscribed materials, the public version excluded evidence window text.

\subsection{Benchmark Composition}

Table~\ref{tab:app-benchmark} summarizes the SR-RAG benchmark composition ($n = 1{,}637$ queries). The benchmark covers 21 common sports rehabilitation sub-conditions, along with two cross-condition guideline sets.

\begin{table}[t]
\centering
\small
\caption{SR-RAG benchmark composition by sub-condition and guideline set.}
\label{tab:app-benchmark}
\setlength{\tabcolsep}{3pt}
\begin{tabular}{llr}
\toprule
\textbf{Code} & \textbf{Sub-condition / Guideline set} & \textbf{$n$} \\
\midrule
\multicolumn{3}{l}{\textbf{Guideline sets}} \\
GL-GEN & General clinical exercise guideline set & 92 \\
GL-SP & Special-population guideline subset & 11 \\
\midrule
\multicolumn{3}{l}{\textbf{Sub-conditions}} \\
ACL & Anterior cruciate ligament injury & 46 \\
AT & Achilles tendinopathy & 36 \\
BSI & Bone stress injury & 38 \\
FS & Adhesive capsulitis of the shoulder & 79 \\
GPA & Groin pain in athletes & 79 \\
GTPS & Greater trochanteric pain syndrome & 96 \\
HSI & Hamstring strain injury & 70 \\
IS & Isthmic spondylolisthesis & 105 \\
LAS & Lateral ankle sprain & 86 \\
LBP & Low back pain & 50 \\
ITBS & Iliotibial band syndrome & 98 \\
LET & Lateral elbow tendinopathy & 70 \\
MACL & Meniscal and articular cartilage lesions & 78 \\
MTSS & Medial tibial stress syndrome & 55 \\
NAHJP & Nonarthritic hip joint pain & 73 \\
NP & Neck pain & 63 \\
PHP & Plantar heel pain & 76 \\
PFP & Patellofemoral pain & 87 \\
PT & Patellar tendinopathy & 77 \\
RCRSP & Rotator cuff related shoulder pain & 102 \\
FTASD & First-time anterior shoulder dislocation & 70 \\
\bottomrule
\end{tabular}
\end{table}

\section{Evaluation Details}
\setcounter{figure}{0}\setcounter{table}{0}\setcounter{equation}{0}\setcounter{algorithm}{0}
\label{app:eval}

Nugget coverage and PICOT match accuracy were implemented via LLM-as-judge \citep{zheng_llm_judge_2023}: for each nugget, the LLM determined whether it was covered by the system answer; for each extracted P/I/C/O/T field, the LLM evaluated whether it matched the reference (paraphrasing permitted; null GT fields were excluded from scoring). Coverage and match rates were then aggregated. Answer faithfulness and semantic similarity were assessed via the RAGAS framework \citep{es_ragas_2024}. RAGAS first extracted claims from the answer, then determined for each claim whether it was supported by the actually retrieved context, yielding a faithfulness score (also LLM-as-judge based). Semantic similarity was computed as cosine similarity between text embeddings of the system answer and GT.

For manual evaluation, review materials included task instructions, scoring criteria, evaluation dimensions, and the question set. For each question, materials provided the question and GT, system-generated answer, retrieved evidence fragments, and a scoring sheet. Detailed scoring criteria for each dimension:
\begin{itemize}
\item \textbf{Medical factual accuracy}: whether conclusions, values, and prescriptions align with current evidence, guidelines, and consensus.
\item \textbf{Answer faithfulness}: whether key statements are substantiated by retrieved fragments.
\item \textbf{Answer relevance}: whether the response addresses core concerns and covers key constraints.
\item \textbf{Safety}: whether the answer contains potentially harmful or misleading content.
\item \textbf{PICOT alignment}: whether the answer is structured around the question's P/I/C/O/T elements without population or intervention mismatch.
\end{itemize}
We aggregated scores from all five experts across five dimensions for the 20 questions and computed means and standard deviations per question and dimension for analysis.

\section{Prompt Templates}
\setcounter{figure}{0}\setcounter{table}{0}\setcounter{equation}{0}\setcounter{algorithm}{0}
\label{app:prompts}

Figure~\ref{lst:prompt-hyde} shows the PICO-guided HyDE prompt, the key generation template in SR-RAG.

\begin{figure}[ht]
\small
\begin{lstlisting}
You are a sports rehabilitation clinician.
Write a concise, evidence-style passage that answers the question, explicitly following PICOT:
- P: population characteristics and key constraints
- I: intervention details (dose/frequency/intensity/progression)
- C: comparator if applicable
- O: outcomes and relevant measures
- T: time horizon / follow-up

Constraints:
- Do NOT cite sources.
- Use neutral clinical language.
- 5-8 sentences, concise, declarative, paper-like.

Question:
{question}

Output ONLY the hypothetical evidence window.
\end{lstlisting}
\caption{Prompt for PICO-guided HyDE generation.}
\label{lst:prompt-hyde}
\end{figure}

\section{Implementation Details}
\setcounter{figure}{0}\setcounter{table}{0}\setcounter{equation}{0}\setcounter{algorithm}{0}
\label{app:impl}

All experiments were implemented in Python 3.11. Graph construction followed Youtu-GraphRAG's schema-bounded extraction and community compression using NetworkX, with graph artifacts stored as JSON. Dense retrieval was implemented with sentence-transformers and FAISS; candidate lists from dense, graph, and HyDE retrieval were fused via RRF. Two-stage reranking used ColBERT (mxbai-edge-colbert-v0) for coarse ranking and a cross-encoder (bge-reranker-v2-m3) for final logits. BETR training used PyTorch 2.6.0 with the Adam optimizer (learning rate: 0.05, epochs: 80, $K=20$ negatives per positive). Hyperparameters ($\tau$) were selected via grid search on the validation split ($n$=327). Unless otherwise specified, LLM decoding used temperature = 0.

Table~\ref{tab:app-hparams} summarizes selected hyperparameters for SR-RAG and BETR.

\begin{table}[t]
\centering
\small
\caption{Selected hyperparameters and training settings.}
\label{tab:app-hparams}
\setlength{\tabcolsep}{4pt}
\begin{tabular}{@{}p{0.44\columnwidth}p{0.44\columnwidth}@{}}
\toprule
\textbf{Item} & \textbf{Setting} \\
\midrule
\multicolumn{2}{l}{\textbf{SR-RAG retrieval and reranking}} \\
Final evidence budget & Top-$K$ = 12 \\
Recall capacities & dense 300; graph 120; HyDE 300; RRF $k$=60 \\
HyDE & 3 passages/query; temp = 0.3 \\
Window selection & 320 tokens; max 3/chunk; overlap 64 \\
Two-stage reranking & ColBERT $\rightarrow$ cross-encoder \\
\midrule
\multicolumn{2}{l}{\textbf{BETR}} \\
Data split & train/val/test: 983/327/327 queries \\
Pair construction & 20 negatives per positive \\
Optimization & Adam, lr 0.05, 80 epochs \\
Shrinkage $\tau$ & grid search; selected $\tau$=1.0 \\
Scale prior $\sigma_a$ & 5.0 \\
\bottomrule
\end{tabular}
\end{table}

\section{Case Study}
\setcounter{figure}{0}\setcounter{table}{0}\setcounter{equation}{0}\setcounter{algorithm}{0}
\label{app:case}

\begin{figure*}[t]
\centering
\includegraphics[width=\textwidth]{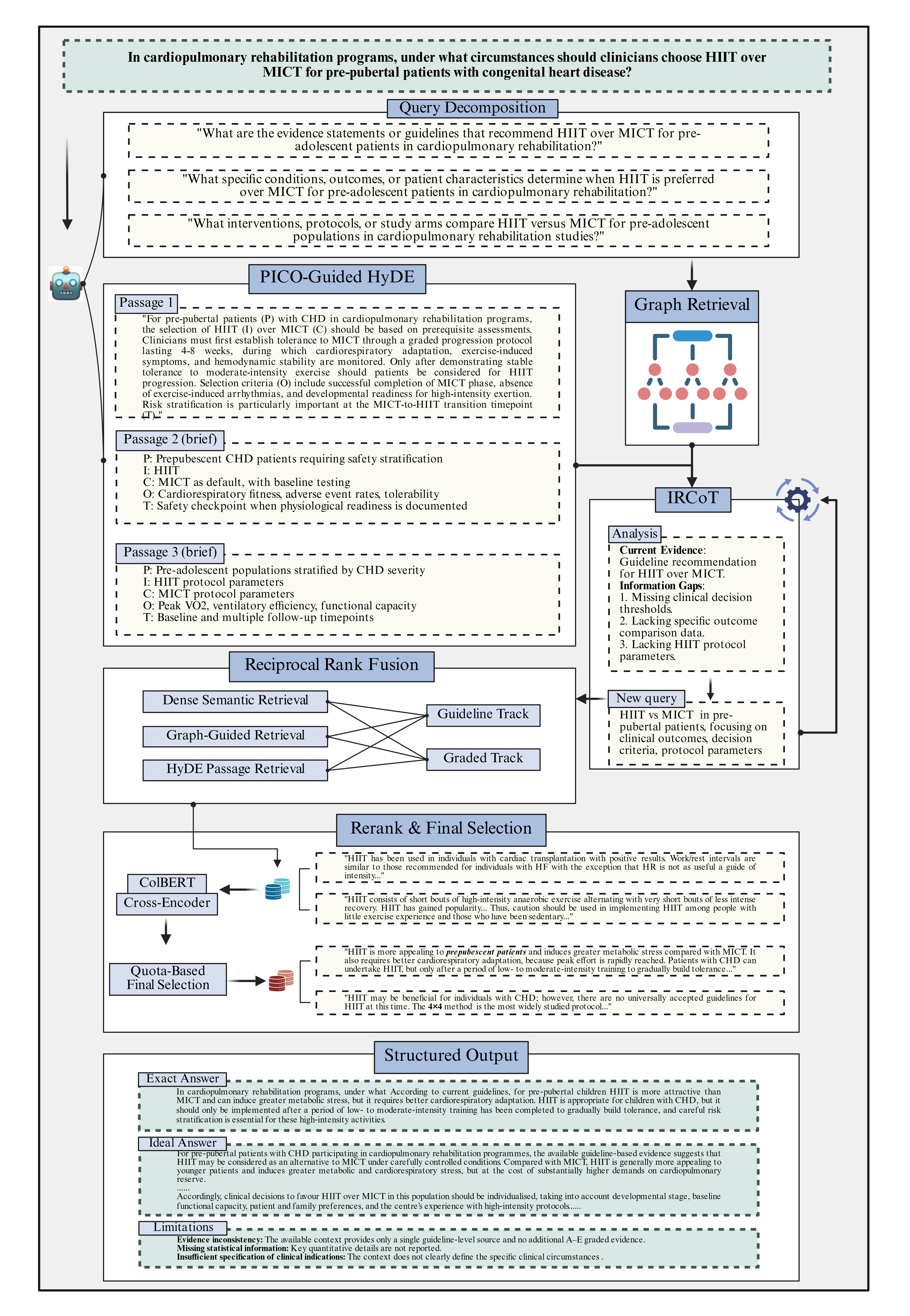}
\caption{Case-study workflow of SR-RAG, illustrating the full pipeline from query to structured output.}
\label{fig:case-study}
\end{figure*}

We present a representative case to illustrate how SR-RAG integrates EBM principles throughout the pipeline. The LLM used was DeepSeek-V3. Figure~\ref{fig:case-study} provides an overview of the case study workflow.

\paragraph{Query.} ``In cardiopulmonary rehabilitation programs, under what circumstances should clinicians choose high-intensity interval training (HIIT) over moderate-intensity continuous training (MICT) for prepubertal patients with congenital heart disease?'' This question exhibits a clear PICO structure and requires cross-evidence integration and conditional reasoning.

\paragraph{PICO-guided HyDE.} The model generated three hypothetical documents with PICO soft constraints (Figure~\ref{fig:case-study}), converting the query into declarative hypothetical answers. The model preferentially reused extractable P/I/C/O/T anchors from the query, tolerating missing fields but prohibiting hallucination of values for absent fields. This bridges the semantic gap between the query and ground truth evidence.

\paragraph{EBM-adaptive Retrieval.} SR-RAG ran separate candidate generation on the Grade A corpus and Grades B--E corpus with their own recall quotas, then merged the candidate sets. BETR calibration was applied to the merged candidates, ensuring that while semantic relevance remains dominant, higher grades were prioritized when relevance is comparable.

\paragraph{Three-channel Fusion and Iterative Reasoning.} Three retrieval channels (Dense, Graph, HyDE) were fused via RRF. The IRCoT module performed three iterations of schema-guided reasoning and reflection (Figure~\ref{fig:case-study}).

\paragraph{Two-stage Reranking and Selection.} ColBERT applied the MaxSim mechanism for coarse ranking; the cross-encoder performed fine-grained ranking within top-$K$. BETR calibration produced the final ranking score $r(q,d)=\hat{a}\,s(q,d)+\hat{u}_{\mathrm{Grade}(d)}$. In this case, two key guideline windows entered the top-2, replacing candidates with potential population mismatch risks (Figure~\ref{fig:case-study}).

\paragraph{Structured Output.} The first paragraph concisely summarizes the core answer; the second elaborates on guidelines and supporting evidence; the third compares the query and response, highlighting limitations (Figure~\ref{fig:case-study}).

\paragraph{Observations.} Three key findings emerge from this case: (i) PICO-guided HyDE successfully extracted anchors (Population: prepubertal CHD patients; Intervention: HIIT; Comparator: MICT; Outcome: cardiopulmonary function) without fabricating missing fields, reducing retrieval drift. (ii) The dual-track retrieval strategy successfully recalled relevant guideline evidence that would have been diluted in a single-pool retrieval, demonstrating the value of evidence-grade-aware candidate generation. (iii) BETR calibration promoted a Grade A guideline window over a Grade C RCT with slightly higher semantic similarity, aligning the final ranking with EBM principles.

\end{document}